\documentclass[10pt,twocolumn,letterpaper]{article}

\usepackage{iccv}
\usepackage{times}
\usepackage{epsfig}
\usepackage{graphicx}
\usepackage{amsmath}
\usepackage{amssymb}
\usepackage{pifont}
\usepackage{enumitem}

\usepackage[pagebackref=true,breaklinks=true,letterpaper=true,colorlinks,bookmarks=false]{hyperref}
\usepackage{color, colortbl}
\usepackage{color}
\definecolor{turquoise}{cmyk}{0.65,0,0.1,0.3}
\definecolor{purple}{rgb}{0.65,0,0.65}
\definecolor{dark_green}{rgb}{0, 0.5, 0}
\definecolor{orange}{rgb}{0.8, 0.6, 0.2}
\definecolor{red}{rgb}{0.8, 0.2, 0.2}
\definecolor{darkred}{rgb}{0.6, 0.1, 0.05}
\definecolor{blueish}{rgb}{0.0, 0.3, .6}
\definecolor{light_gray}{rgb}{0.7, 0.7, .7}
\definecolor{pink}{rgb}{1, 0, 1}
\definecolor{greyblue}{rgb}{0.25, 0.25, 1}

\newcommand{\FIDm}{$\text{FID}_{g}$}
\newcommand{\FIDmArrow}{$\text{FID}_{g}\downarrow$}
\newcommand{\FIDk}{$\text{FID}_{k}$}
\newcommand{\FIDkArrow}{$\text{FID}_{k}\downarrow$}
\newcommand{\Divm}{$\text{Dist}_{m}$}
\newcommand{\DivmArrow}{$\text{Dist}_{g}\uparrow$}
\newcommand{\Divk}{$\text{Dist}_{k}$}
\newcommand{\DivkArrow}{$\text{Dist}_{k}\uparrow$}

\newcommand{\BeatMetricArrow}{$\text{BeatAlign}\uparrow$}

\newcommand{\cmark}{\ding{51}}
\newcommand{\xmark}{\ding{55}}

\newcommand{\RealMotion}{\mathbb{R}^{T\times N\times 3}}

\newcommand\blfootnote[1]{%
  \begingroup
  \renewcommand\thefootnote{}\footnote{#1}%
  \addtocounter{footnote}{-1}%
  \endgroup
}

\usepackage{booktabs,multirow,array,tabularx}
\usepackage[table, dvipsnames]{xcolor}
\definecolor{Gray}{gray}{0.95}
\usepackage{caption}
    \newcommand{\midsepremove}{\aboverulesep = 0mm \belowrulesep = 0mm}
    \midsepremove
    \newcommand{\midsepdefault}{\aboverulesep = 0.605mm \belowrulesep = 0.984mm}
    \midsepdefault
    \newcommand{\tabincell}[2]{\begin{tabular}{@{}#1@{}}#2\end{tabular}}  
\usepackage{booktabs}

\usepackage[breaklinks=true,bookmarks=false]{hyperref}

\iccvfinalcopy 

\def\iccvPaperID{8465} 

\ificcvfinal\fi

\begin{document}

\title{AI Choreographer: Music Conditioned 3D Dance Generation with AIST++}

\author{Ruilong Li$^{*1}$\qquad Shan Yang$^{*2}$\qquad David A. Ross$^{2}$ \qquad Angjoo Kanazawa$^{2,3}$\\
$^1$University of Southern California\qquad $^2$Google Research \qquad $^3$University of California, Berkeley
}

\ificcvfinal\thispagestyle{empty}\fi

\twocolumn[
\begin{@twocolumnfalse}{%
\maketitle
\vspace{-5mm}
\centering
    \includegraphics[width=0.95\textwidth]{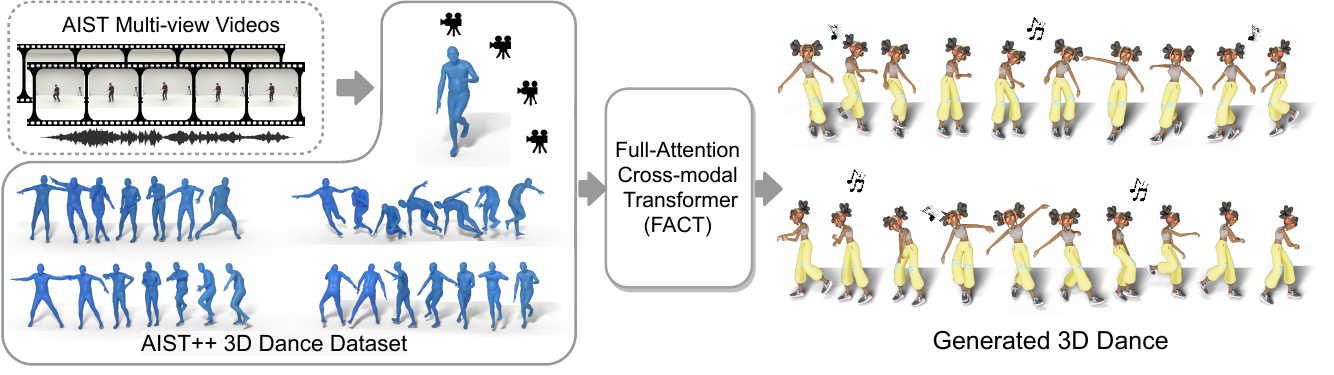}
    \captionof{figure}{
      \textbf{AI Choreographer.} 
      We present a new 3D dance dataset, AIST++, which contains $5.2$ hours of 3D motion reconstructed from real dancers paired with music (left) and a novel Full-Attention Cross-modal Transformer (FACT) network that can generate realistic 3D dance motion with global translation conditioned on music (right). We output our 3D motion in representations that allow for instant motion retargeting to a novel character. Here we use a character from Mixamo \cite{mixamo}
    }
    \label{fig:teaser}
    \vspace{3mm}
}\end{@twocolumnfalse}
]
\begin{abstract}
We present AIST++, a new multi-modal dataset of 3D dance motion and music,  along with FACT, a Full-Attention Cross-modal Transformer network for generating 3D dance motion conditioned on music. 
The proposed AIST++ dataset contains 5.2 hours of 3D dance motion in 1408 sequences, covering 10 dance genres with multi-view videos with known camera poses---the largest dataset of this kind to our knowledge. We show that naively applying sequence models such as transformers to this dataset for the task of music conditioned 3D motion generation does not produce satisfactory 3D motion that is well correlated with the input music. We overcome these shortcomings by introducing key changes in its architecture design and supervision: FACT model involves a deep cross-modal transformer block with full-attention that is trained to predict $N$ future motions. We empirically show that these changes are key factors in generating long sequences of realistic dance motion that are well-attuned to the input music. 
We conduct extensive experiments on AIST++ with user studies, where our method outperforms recent state-of-the-art methods both qualitatively and quantitatively. 
The code and the dataset can be found at: \href{https://google.github.io/aichoreographer}{https://google.github.io/aichoreographer}. %
\blfootnote{ 
$^*$ equal contribution. Work performed while Ruilong was an intern at Google. 
}
\end{abstract}

\section{Introduction}
\begin{figure*}[t]
\centering
\vspace{-6mm}
\includegraphics[width=0.95\textwidth]{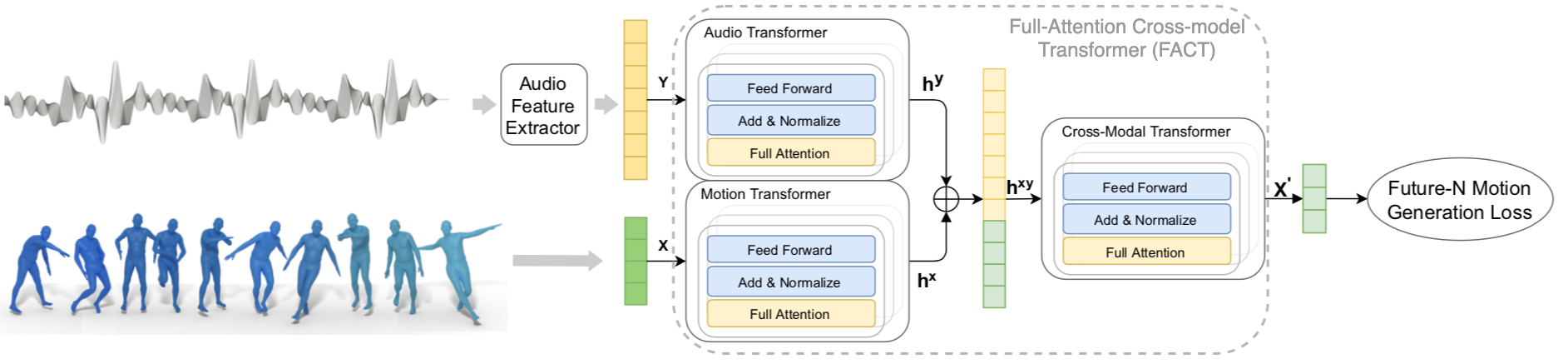}
\vspace{-3mm}
\caption{\textbf{Cross-Modal Music Conditioned 3D Motion Generation Overview.} Our proposed a Full-Attention Cross-modal Transformer (FACT) network (details in Figure~\ref{fig:attn_comparison}) takes in a music piece and a $2$-second sequence of seed motion, then auto-regressively generates long-range future motions that correlates with the input music. }
\label{fig:model_overview}
\vspace{-3mm}
\end{figure*}
The ability to dance by composing movement patterns that align to musical beats is a fundamental aspect of human behavior.
Dancing is an universal language found in all cultures \cite{todanceishuman}, and today, many people express themselves through dance on contemporary online media platforms. The most watched videos on YouTube are dance-centric music videos such as ``Baby Shark Dance'', and ``Gangnam Style''~\cite{TopYTVideos}, making 
dance a more and more powerful tool to spread messages across the internet. 
However, dancing is a form of art that requires practice---even for humans, professional training is required to equip a dancer with a rich repertoire of dance motions to create an expressive choreography. Computationally,
this is even more challenging as the task requires the ability to generate a continuous motion with high kinematic complexity that captures the non-linear relationship with the accompanying music. 

In this work, we address these challenges by presenting a novel Full Attention Cross-modal Transformer (FACT) network, which can robustly generate realistic 3D dance motion from music, along with a large-scale multi-modal 3D dance motion dataset, AIST++, to train such a model. Specifically, given a piece of music and a short (2 seconds) seed motion, our model is able to generate a long sequence of realistic 3D dance motions. 
Our model effectively learns the music-motion correlation and can generate dance sequences that varies for different input music. We represent dance as a 3D motion sequence that consists of joint rotation and global translation, which enables easy transfer of our output for applications such as motion retargeting as shown in Figure~\ref{fig:teaser}. 

In order to generate 3D dance motion from music, we propose a novel Full Attention Cross-modal Transformer (FACT) model,
which employs an audio transformer and seed motion transformer to encode the inputs, which are then fused by a cross-modal transformer that models the distribution between audio and motion. This model is trained to predict $N$ future motion sequences and at test time is applied in an auto-regressive manner to generate continuous motion. 
The success of our model relies on three key design choices: 1)~the use of full-attention in an auto-regressive model, 2)~future-N supervision, and 3)~early fusion of two modalities. The combination of these choices is critical for training a model that can generate a long realistic dance motion that is attuned to the music.
Although prior work has explored using transformers for motion generation~\cite{aksan2020attention}, we find that naively applying transformers to the 3D dance generation problem without these key choices does not lead to a very effective model.  

In particular, we notice that because the context window in the motion domain is significantly smaller than that of language models, it is possible to apply full-attention transformers in an auto-regressive manner, which leads to a more powerful model. It is also critical that the full-attention transformer is trained to predict $N$ possible future motions instead of one. These two design choices are key for preventing 3D motion from freezing or drifting after several auto-regressive steps as reported in prior works on 3D motion generation~\cite{aksan2019structured, aksan2020attention}. Our model is trained to predict 20 future frames, but it is able to produce realistic 3D dance motion for over 1200 frames at test time. We also show that fusing the two modalities early, resulting in a deep cross-modal transformer, is important for training a model that generates different dance sequences for different music. 

In order to train the proposed model, we also address the problem of data. While there are a few motion capture datasets of dancers dancing to music, collecting mocap data requires heavily instrumented environments making these datasets severely limited in the number of available dance sequences, dancer and music diversity. In this work, we propose a new dataset called AIST++, which we build from the existing multi-view dance video database called AIST~\cite{aist-dance-db}.  We use the multi-view videos to recover reliable 3D motion from this data. We will release code and this dataset for research purposes, where AIST++ can be a new benchmark for the task of 3D dance generation conditioned on music.

In summary, our contributions are as follows:
\begin{itemize}
    \item We propose Full Attention Cross-Modal Transformer model, FACT, which can generate a long sequence of realistic 3D dance motion that is well correlated with the input music. 

    \item We introduce AIST++ dataset containing 5.2 hours of 3D dance motions accompanied with music and multi-view images, which to our knowledge is the largest dataset of such kind. 

    \item We provide extensive evaluations validating our design choices and show that they are critical for high quality, multi-modal, long motion sequence generation.
    
\end{itemize}
\section{Related Work}
\vspace{-1mm}
\paragraph{3D Human Motion Synthesis}
The problem of generating realistic and controllable 3D human motion sequences has long been studied.
Earlier works employ statistical models such as kernel-based probability distribution~~\cite{pullen2000animating,bowden2000learning,galata2001learning,brand2000style} to synthesize motion, but abstract away motion details.
Motion graphs~\cite{lee1999hierarchical,arikan2002interactive,kovar2008motion} address this problem by generating motions in a non-parametric manner.
Motion graph is a directed graph constructed on a corpus of motion capture data, where each node is a pose and the edges represent the transition between poses. Motion is generated by a random walk on this graph.
A challenge in motion graph is in generating plausible transition that some approaches address via parameterizing the transition~\cite{heck2007parametric}. 
With the development in deep learning, many approaches explore the applicability of neural networks to generate 3D motion by training on a large-scale motion capture dataset, where 
network architectures such as CNNs~\cite{holden2015learning, holden2016deep}, GANs~\cite{hernandez2019human}, RBMs~\cite{taylor2009factored}, RNNs~\cite{fragkiadaki2015recurrent,aksan2019structured,jain2016structural,ghosh2017learning,chiu2019action,du2019bio,wang2019imitation,butepage2017deep,villegas2018neural} and Transformers~\cite{aksan2020attention,bhattacharya2021text2gestures} have been explored. 
Auto-regressive models like RNNs and vanilla Transformers are capable of generating unbounded motion in theory, but in practice suffer from regression to the mean where motion ``freezes'' after several iterations, or drift to unnatural motions~\cite{aksan2019structured, aksan2020attention}. Some works~\cite{bengio2015scheduled, li2018auto, kundu2020cross} propose to ease this problem by periodically using the network’s own outputs as inputs during training.
Phase-functioned neural networks and it's variations~\cite{zhang2018mode, holden2017phase, starke2019neural, starke2020local}  address this issue via conditioning the network weights on phase, however, they do not scale well to represent a wide variety of motion. 

\vspace{-3mm}
\paragraph{Audio To Human Motion Generation}
Audio to motion generation has been studied in 2D pose context either in optimization based approach~\cite{tendulkar2020feel}, or learning based approaches~\cite{lee2018listen,shlizerman2018audio,lee2019dance,ren2019music, ren2020self, ferreira2020learning}
where 2D pose skeletons are generated from a conditioning audio. Training data for 2D pose and audio is abundant thanks to the high reliability of 2D pose detectors~\cite{cao2018openpose}. 
However, predicting motion in 2D is limited in its expressiveness and potential for downstream applications.
For 3D dance generation, earlier approaches explore matching existing 3D motion to music~\cite{shiratori2006dancing} using motion graph based approach~\cite{fan2011example}. More recent approach employ LSTMs~\cite{alemi2017groovenet,tang2018dance,yalta2019weakly,zhuang2020towards,kao2020temporally}, GANs~\cite{lee2019dance,sun2020deepdance,ginosar2019learning}, transformer encoder with RNN decoder~\cite{huang2021} or convolutional~\cite{ahn2020generative, ye2020choreonet} sequence-to-sequence models.
Concurrent to our work, Chen~\etal\cite{chen2021choreomaster} proposed a method that is based on motion graphs with learned embedding space.
Many prior works~\cite{shlizerman2018audio,ren2020self,kao2020temporally,ginosar2019learning,ye2020choreonet} solve this problem by predicting future motion deterministically from audio without seed motion. When the same audio has multiple corresponding motions, which often occurs in dance data, these methods collapse to predicting a mean pose. 
In contrast, we formulate the problem with seed motion as in~\cite{li2020learning, zhuang2020music2dance}, which allows generation of multiple motion from the same audio even with a deterministic model. 

Closest to our work is that of Li~\etal~\cite{li2020learning}, which also employ transformer based architecture but only on audio and motion. Furthermore, their approach discretize the output joint space in order to account for multi-modality, which generates unrealistic motion. In this work we introduce a novel full-attention based cross-modal transformer (FACT model) for audio and motion, which can not only preserve the correlation between music and 3D motion better, but also generate more realistic long 3D human motion with global translation. 
One of the biggest bottleneck in 3D dance generation approaches is that of data. Recent work of Li~\etal~\cite{li2020learning} reconstruct 3D motion from dance videos on the Internet, however the data is not public. Further, using 3D motion reconstructed from monocular videos may not be reliable and lack accurate global 3D translation information. 
In this work we also reconstruct the 3D motion from 2D dance video, but from multi-view video sequences, which addresses these issues. While there are many large scale 3D motion capture datasets~\cite{h36m_pami,mahmood2019amass, mixamo, DIP:SIGGRAPHAsia:2018}, mocap dataset of 3D dance is quite limited as it requires heavy instrumentation and expert dancers for capture. As such, many of these previous works operate on either small-scale or private motion capture datasets~\cite{tang2018dance, alemi2017groovenet, zhuang2020music2dance}. We compare our proposed dataset with these public datasets in Table~\ref{tab:dataset_comparison}.

\vspace{-3mm}
\paragraph{Cross-Modal Sequence-to-Sequence Generation}
Beyond of the scope of human motion generation, our work is closely related to the research of using neural network on cross-modal sequence to sequence generation task. In natural language processing and computer vision, tasks like text to speech (TTS)~\cite{ren2019fastspeech,jia2019leveraging,karita2019comparative,valin2019lpcnet} and speech to gesture~\cite{ferstl2018investigating,ginosar2019learning,ferstl2020adversarial}, image/video captioning (pixels to text)~\cite{bybaby,karpathy2015deep,lu2018neural,krishna2017dense} involve solving the cross-modal sequence to sequence generation problem. Initially, combination of CNNs and RNNs~\cite{venugopalan2014translating,venugopalan2015sequence,yao2015describing,yu2016video} were prominent in approaching this problem.
More recently, with the development of attention mechanism~\cite{vaswani2017attention}, transformer based networks achieve top performance for visual-text~\cite{zhou2018end,sun2019videobert,duan2018weakly,li2019entangled,iashin2020multi,sun2019learning,sun2019learning}, visual-audio~\cite{gan2020foley,xu2020cross} cross-modal sequence to sequence generation task. Our work explores audio to 3D motion in a transformer based architecture.  
While all cross-modal problems induce its own challenges, the problem of music to 3D dance is uniquely challenging in that there are many ways to dance to the same music and that the same dance choreography may be used for multiple music. We hope the proposed AIST++ dataset advances research in this relatively under-explored problem. 

\section{AIST++ Dataset}
\begin{table*}[t]
\large
\vspace{-4mm}
\centering{
\resizebox{1.0\linewidth}{!}{%
\midsepremove
\begin{tabular}{m{4.2cm}cccccccccc}
\toprule
Dataset & Music & 3D $\text{Joint}_{\text{pos}}$ & 3D $\text{Joint}_{\text{rot}}$ & 2D Kpt  & Views & Images & Genres & Subjects & Sequences & Seconds \\
\hline
AMASS\cite{mahmood2019amass} & \xmark & \cmark & \cmark & \xmark & 0 & 0 & 0 & 344 & 11265 & 145251 \\
Human3.6M\cite{h36m_pami} & \xmark & \cmark & \cmark & \cmark & 4 & 3.6M & 0 & 11 & 210 & 71561 \\
\hline
Dance with Melody\cite{tang2018dance} & \cmark & \cmark & \xmark & \xmark & 0 & 0 & 4 & - & 61 & 5640 \\
GrooveNet~\cite{alemi2017groovenet}& \cmark & \cmark & \xmark & \xmark & 0 & 0 & 1 & 1 & 2 & 1380 \\
DanceNet~\cite{zhuang2020music2dance} & \cmark & \cmark & \xmark & \xmark & 0 & 0 & 2 & 2 & 2 & 3472 \\
EA-MUD~\cite{sun2020deepdance} & \cmark & \cmark & \xmark & \xmark & 0 & 0 & 4 & - & 17 & 1254 \\
\hline
\rowcolor{Gray}
AIST++ & \cmark & \cmark & \cmark & \cmark& 9 & 10.1M & 10 & 30 & 1408 & 18694 \\
\bottomrule
\end{tabular}}}
\midsepdefault
\vspace{2mm}
\caption{\textbf{3D Dance Datasets Comparisons.} The proposed AIST++ dataset is the largest dataset with 3D dance motion paired with music. We also have the largest variety of subjects and genres. Furthermore, our dataset is the only one that comes with image frames, as other dance datasets only contain motion capture dataset. We include popular 3D motion dataset without any music in the first two rows for reference. 
}
\label{tab:dataset_comparison}
\vspace{-3mm}
\end{table*}
\noindent\textbf{Data Collection}
We generate the proposed 3D motion dataset from an existing database called AIST Dance Database~\cite{aist-dance-db}. AIST is only a collection of videos without any 3D information. Although it contains multi-view videos of dancers, these cameras are not calibrated, making 3D reconstruction of dancers a non-trivial effort. We recover the camera calibration parameters and the 3D human motion in terms of SMPL parameters. Please find the details of this algorithm in the Appendix. Although we adopt the best practices in reconstructing this data, no code base exist for this particular problem setup and running this pipeline on a large-scale video dataset requires non-trivial amount of compute and effort. We will make the 3D data and camera parameters publicly available, which allows the community to benchmark on this dataset on an equal footing. 

\paragraph{Dataset Description}
Resulting AIST++ is a large-scale 3D human dance motion dataset that contains a wide variety of 3D motion paired with music. It has the following extra annotations for each frame:
\begin{itemize}
    \setlength\itemsep{0.1em}
    \item $9$ views of camera intrinsic and extrinsic parameters;
    \item $17$ COCO-format\cite{Ronchi_2017_ICCV} human joint locations in both 2D and 3D;
    \item $24$ SMPL~\cite{SMPL} pose parameters along with the global scaling and translation.
\end{itemize}
Besides the above properties, AIST++ dataset also contains multi-view synchronized image data unlike prior 3D dance dataset, making it useful for other research directions such as 2D/3D pose estimation.
To our knowledge, AIST++ is the largest 3D human dance dataset
with $\mathbf{1408}$ sequences, $\mathbf{30}$ subjects and $\mathbf{10}$ dance genres with basic and advanced choreographies. See Table.~\ref{tab:dataset_comparison} for comparison with other 3D motion and dance datasets. AIST++ is a complementary dataset to existing 3D motion dataset such as AMASS~\cite{mahmood2019amass}, which contains only $17.8$ minutes of dance motions with no accompanying music. 

Owing to the richness of AIST, AIST++ contains 10 dance genres: Old School (Break, Pop, Lock and Waack) and New School (Middle Hip-hop, LA-style Hip-hop, House, Krump, Street Jazz and Ballet Jazz). Please see the Appendix for more details and statistics. 
The motions are equally distributed among all dance genres, covering wide variety of music tempos denoted as beat per minute (BPM)\cite{moelants2003dance}. Each genre of dance motions contains $85\%$ of basic choreographies and $15\%$ of advanced choreographies, in which the former ones are those basic short dancing movements while the latter ones are longer movements freely designed by the dancers. However, note that AIST is an instructional database and records multiple dancers dancing the same choreography for different music with varying BPM, a common practice in dance. This posits a unique challenge in cross-modal sequence-to-sequence generation. We carefully construct non-overlapping train and val subsets on AIST++ to make sure neither choreography nor music is shared across the subsets. 

\section{Music Conditioned 3D Dance Generation}
Here we describe our approach towards the problem of music conditioned 3D dance generation. Specifically, given a $2$-second seed sample of motion represented as $\mathbf{X}=(x_1, \dots, x_T)$ and a longer conditioning music sequence represented as $\mathbf{Y}=({y}_1,\dots,{y}_{T'})$, the problem is to generate a sequence of future motion ${\mathbf{X'}}=({x}_{T+1},\dots,{x}_{T'})$ from time step $T+1$ to $T'$, where $T'\gg T$. 

\begin{figure*}[t]
\vspace{-5mm}
\centering
\includegraphics[width=0.97\textwidth]{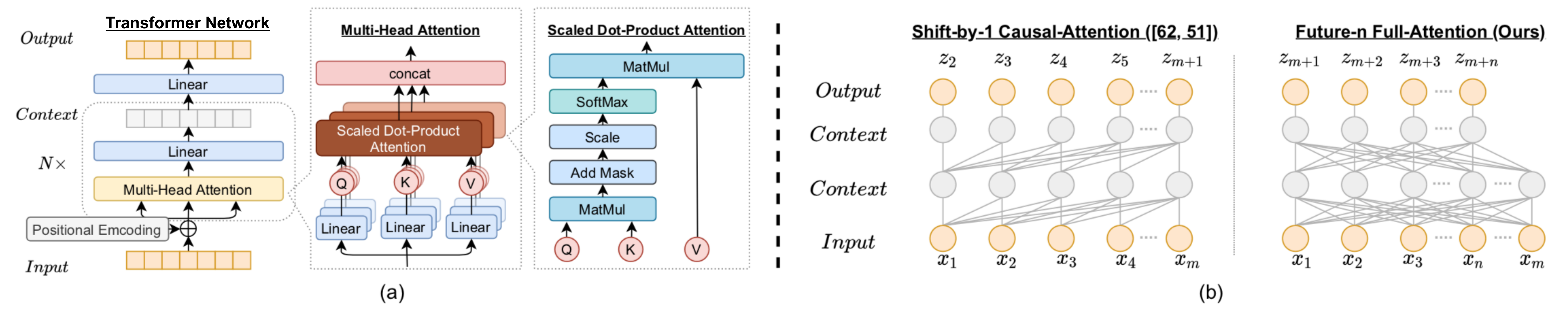}
\vspace{-2mm}
\caption{\textbf{FACT Model Details.} (a) The structure of the audio/motion/cross-modal transformer with $N$ attention layers.
(b) Attention and supervision mechanism as a simplified two-layer model. Models like GPT~\cite{radford2018improving} and the motion generator of~\cite{li2020learning} use causal attention  (left) to predict the immediate next output for each input nodes. We employ full-attention and predict $n$ future from the last input timestamp $m$ (right). 
The dots on the bottom row are the input tensors, which are computed into context tensors through causal (left) and full (right) attention transformer layer. The output (predictions) are shown on the top. 
We empirically show that these design choices are critical in generating non-freezing, more realistic motion sequences.
}
\label{fig:attn_comparison}
\vspace{-3mm}
\end{figure*}
\vspace{-4mm}
\paragraph{Preliminaries}
Transformer~\cite{vaswani2017attention} is an attention based network widely applied in natural language processing. 
A basic transformer building block (shown in of Figure~\ref{fig:attn_comparison} (a)) has multiple layers with each layer composed of a multi-head attention-layer (Attn) followed by a feed forward layer (FF). 
The multi-head attention-layer embeds input sequence $\mathbf{X}$ into an internal representation often referred to as the context vector $\mathbf{C}$. 
Specifically, the output of the attention layer, the context vector $\mathbf{C}$ is computed using the query vector ${\mathbf{Q}}$ and the key ${\mathbf{K}}$ value  ${\mathbf{V}}$ pair from input with or without a mask ${\mathbf{M}}$ via,
\begin{align}
    {\mathbf{C}}&= \text{FF}(\text{Attn}(\mathbf{Q,K,V, M})) \notag \\
    &=\text{FF}(\text{softmax}\Bigg(\frac{\mathbf{QK}^T+\mathbf{M}}{\sqrt{D}}\Bigg)\mathbf{V}), \notag \\
    \mathbf{Q}&=\mathbf{X}\mathbf{W}^Q, \mathbf{K}=\mathbf{X}\mathbf{W}^K, \mathbf{V}=\mathbf{X}\mathbf{W}^V
    \label{eqn:attn}
\end{align}
where $D$ is the number of channels in the attention layer and $\mathbf{W}$ are trainable weights. The design of the mask function is a key parameter in a transformer.
In natural language generation, causal models such as GPT~\cite{radford2018improving} uses an upper triangular look-ahead mask $\mathbf{M}$ to enable causal attention where each token can only look at past inputs. This allows efficient inference at test time, since intermediate context vectors do not need to be recomputed, especially given the large context window in these models (2048). On the other hand, models like BERT~\cite{devlin2018bert} employ full-attention for feature learning, but rarely are these models employed in an auto-regressive manner, due to its inefficiency at test time.

\subsection{Full Attention Cross-Modal Transformer} \label{sec:cross_modal_transformer}
We propose Full Attention Cross-Modal Transformer (FACT) model for the task of 3D dance motion generation. Given the seed motion ${\mathbf{X}}$ and audio features $\mathbf{Y}$, FACT first encodes these inputs using a motion transformer $f_{\text{mot}}$ and audio transformer $f_{\text{audio}}$ into motion and audio embeddings ${\mathbf{h}^x}_{1:T}$ and ${\mathbf{h}^y}_{1:T'}$ respectively. These are then concatenated and sent to a cross-modal transformer $f_{\text{cross}}$, which learns the correspondence between both modalities and generates $N$ future motion sequences $\mathbf{X'}$, which is used to train the model in a self-supervised manner. All three transformers are jointly learned in an end-to-end manner. This process is illustrated in Figure~\ref{fig:model_overview}. At test time, we apply this model in an auto-regressive framework, where we take the first predicted motion as the input of the next generation step and shift all conditioning by one.

FACT involves three key design choices that are critical for producing realistic 3D dance motion from music. First, all of the transformers use full-attention mask. We can still apply this model efficiently in an auto-regressive framework at test time, since our context window is not prohibitively large (240). The full-attention model is more expressive than the causal model because internal tokens have access to all inputs. Due to this full-attention design, we train our model to only predict the unseen future after the context window. In particular, we train our model to predict $N$ futures beyond the current input instead of just $1$ future motion. This encourages the network to pay more attention to the temporal context, and we experimentally validate that this is a key factor training a model that does not suffer from motion freezing or diverging after a few generation steps. This attention design is in contrast to prior work that employ transformers for the task of 3D motion~\cite{aksan2020attention} or dance generation~\cite{li2020learning}, which applies GPT~\cite{radford2018improving} style causal transformer trained to predict the immediate next future token. We illustrate this difference in Figure~\ref{fig:attn_comparison} (b). 

Lastly, we fuse the two embeddings early and employ a deep 12-layer cross-modal transformer module. This is in contrast to prior work that used a single MLP to combine the audio and motion embeddings~\cite{li2020learning}, and we find that deep cross-modal module is essential for training a model that actually pays attention to the input music. This is particularly important as in dance, similar choreography can be used for multiple music. This also happens in AIST dataset, and we find that without a deep cross-modal module, the network is prone to ignoring the conditioning music. We experimentally validate this in Section~\ref{sec:ablation_study}. 

\section{Experiments}
\label{sec:exp}
\begin{table*}[t]
\vspace{-4mm}
\centering{
{%
\midsepremove
\begin{tabular}[c]{lcccccc}
\toprule
 & \multicolumn{2}{c}{Motion Quality} & \multicolumn{2}{c}{Motion Diversity} & \multicolumn{1}{c}{Motion-Music Corr} & User Study  \\ 
\cmidrule(lr){2-3} \cmidrule(lr){4-5} \cmidrule(lr){6-6} \cmidrule(lr){7-7}
 & \tabincell{c}{\FIDkArrow} & \tabincell{c}{\FIDmArrow} & \tabincell{c}{\DivkArrow} & \tabincell{c}{\DivmArrow} & \tabincell{c}{\BeatMetricArrow} & \tabincell{c}{FACT WinRate$\downarrow$} \\
\hline
AIST++ & -- & -- & 9.057 & 7.556 & 0.292 & --  \\
AIST++ (random)  & -- & -- & -- & -- & 0.213 & 25.4\% \\
\hline
Li~\etal\cite{li2020learning} & 86.43 & 20.58 & 6.85* & 4.93 & 0.232 & 80.6\% \\
Dancenet~\cite{zhuang2020music2dance} & 69.18 & 17.76 & 2.86 & 2.72 & 0.232 & 71.1\%\\
DanceRevolution~\cite{huang2021} & 73.42 & 31.01 & 3.52 & 2.46 & 0.220 & 77.0\% \\
\hline
\rowcolor{Gray}
FACT (ours) & \textbf{35.35} & \textbf{12.40} & \textbf{5.94} & \textbf{5.30}  & \textbf{0.241} & -- \\
\bottomrule
\end{tabular}}}
\midsepdefault
\caption{\textbf{Conditional Motion Generation Evaluation on AIST++ dataset.} {Comparing to the three recent state-of-the-art methods, our model generates motions that are more realistic, better correlated with input music and more diversified when conditioned on different music. }
*Note Li~\etal~\cite{li2020learning}'s generated motions are discontinuous making its average kinetic feature distance (\FIDk) abnormally high. 
}
\label{tab:exp_comparison}
\vspace{-3mm}
\end{table*}

\subsection{AIST++ Motion Quality Validation}
\label{sec:motion_eval}
We first carefully validate the quality of our 3D motion reconstruction. 
Possible error sources that may affect the quality of our 3D reconstruction include inaccurate 2D keypoints detection and the estimated camera parameters.
As there is no 3D ground-truth for AIST dataset, our validation here is based-on the observation that the re-projected 2D keypoints should be consistent with the predicted 2D keypoints which have high prediction confidence in each image. 
We use the 2D mean per joint position error MPJPE-2D, commonly used for 3D reconstruction quality measurement~\cite{kocabas2019vibe, h36m_pami, multiviewpose}) to evaluate the consistency between the predicted 2D keypoints and the reconstructed 3D keypoints along with the estimated camera parameters. 
Note we only consider 2D keypoints with prediction confidence over 0.5 to avoid noise. 
The $\text{MPJPE-2D}$ of our entire dataset is $6.2$ pixels on the $1920\times1080$ image resolution, and over $86\%$ of those has less than $10$ pixels of error. 
Besides, we also calculate the PCKh metric introduced in~\cite{andriluka20142d} on our AIST++. The PCKh@0.5 on the whole set is $98.7\%$, meaning the reconstructed 3D keypoints are highly consistent with the predicted 2D keypoints. 
Please refer to the Appendix for detailed analysis of MPJPE-2D and PCKh on AIST++.

\subsection{Music Conditioned 3D Motion Generation}

\subsubsection{Experimental Setup}
\paragraph{Dataset Split}
All the experiments in this paper are conducted on our AIST++ dataset, which to our knowledge is the largest dataset of this kind. 
We split AIST++ into \emph{train} and \emph{test} set, and report the performance on the \emph{test} set only. We carefully split the dataset to make sure that the music and dance motion in the \emph{test} set does not overlap with that in the \emph{train} set. 
To build the \emph{test} set, we first select one music piece from each of the 10 genres. Then for each music piece, we randomly select two dancers, each with two different choreographies paired with that music, resulting in total $40$ unique choreographies in the \emph{test} set. The \emph{train} set is built by excluding all test musics and test choreographies from AIST++, resulting in total $329$ unique choreographies in the \emph{train} set. Note that in the test set we \textit{intentionally} pick music pieces with different BPMs so that it covers all kinds of BPMs ranging from $80$ to $135$ in AIST++.  

\vspace{-3mm}
\paragraph{Implementation Details}
In our main experiment, the input of the model contains a seed motion sequence with $120$ frames (2 seconds) and a music sequence with $240$ frames (4 seconds), where the two sequences are aligned on the first frame. The output of the model is the future motion sequence with $N=20$ frames supervised by $L2$ loss.
During inference we continually generate future motions in a auto-regressive manner at $60$ FPS, where only the first predicted motion is kept in every step.
We use the publicly available audio processing toolbox Librosa~\cite{mcfee2015librosa} to extract the music features including: 1-dim \emph{envelope}, 20-dim \emph{MFCC}, 12-dim \emph{chroma}, 1-dim \emph{one-hot peaks} and 1-dim \emph{one-hot beats}, 
resulting in a 35-dim music feature. 
We combine the 9-dim rotation matrix representation for all $24$ joints, along with a 3-dim global translation vector, resulting in a 219-dim motion feature. Both these raw audio and motion features are first embedded into $800$-dim hidden representations with linear layers, then added with learnable positional encoding, before they were input into the transformer layers.
All the three (audio, motion, cross-modal) transformers have $10$ attention heads with $800$ hidden size. The number of attention layers in each transformer varies based on the experiments, as described in Sec.~\ref{sec:ablation_study}. 
We disregard the last linear layer in the audio/motion transformer and the positional encoding in the cross-modal transformer, as they are not necessary in the FACT model.
All our experiments are trained with $16$ batch size using Adam~\cite{kingma2014adam} optimizer. 
The learning rate starts from $1\mathrm{e}{-4}$ and drops to \{$1\mathrm{e}{-5}$, $1\mathrm{e}{-6}$\} after \{$60k$, $100k$\} steps. The training finishes after $300k$, which takes $3$ days on $4$ TPUs. 
For baselines, we compare with the latest work on 3D dance generation that take music and seed motion as input,
including Dancenet~\cite{zhuang2020music2dance} and Li~\etal~\cite{li2020learning}. For a more comprehensive evaluation we also compare with the recent state-of-the-art 2D dance generation method DanceRevolution~\cite{huang2021}. We adapt this work to output 3D joint locations which can be directly compared with our results quantitatively, though joint locations do not allow immediate re-targeting.
We train and test these baselines on the same dataset with ours using the \emph{official} code provided by the authors.

\subsubsection{Quantitative Evaluation}
In this section, we evaluate our proposed model FACT on the following aspects: (1) motion quality, (2) generation diversity and (3) motion-music correlation. 
Experiments results (shown in Table~\ref{tab:exp_comparison}) show that our model out-performs state-of-the-art methods~\cite{li2020learning, huang2021, zhuang2020music2dance}, on those criteria.

\paragraph{Motion Quality} 
Similar to prior works~\cite{li2020learning, huang2021}, we evaluate the generated motion quality by calculating the distribution distance between the generated and the ground-truth motions using Frechet Inception Distance (FID)~\cite{heusel2017gans} on the extracted motion features. {As prior work used motion-encoders that are not public, we measure FID with two well-designed motion feature extractors~\cite{muller2005efficient, onuma2008fmdistance} implemented in \emph{fairmotion}~\cite{gopinath2020fairmotion}: (1) a geometric feature extractor that produces a boolean vector $\mathbf{z}_{g} \in \mathbb{R}^{33}$ expressing geometric relations between certain body points in the motion sequence $X \in \RealMotion$, (2) a kinetic feature extractor~\cite{onuma2008fmdistance} that maps a motion sequence $X$ to $\mathbf{z}_{k} \in \mathbb{R}^{72}$, which represents the kinetic aspects of the motion such as velocity and accelerations. We denote the FID based on these geometric and kinetic features as \FIDm and \FIDk, respectively.
The metrics are calculated between the real dance motion sequences in AIST++ test set and $40$ generated motion sequences each with $T=1200$ frames (20 secs). }
As shown in Table~\ref{tab:exp_comparison}, our generated motion sequences have a much closer distribution to ground-truth motions compared with the three baselines.
We also visualize the generated sequences from the baselines in our supplemental video.
\begin{figure}[t]
\vspace{-8mm}
\centering
\includegraphics[width=0.46\textwidth]{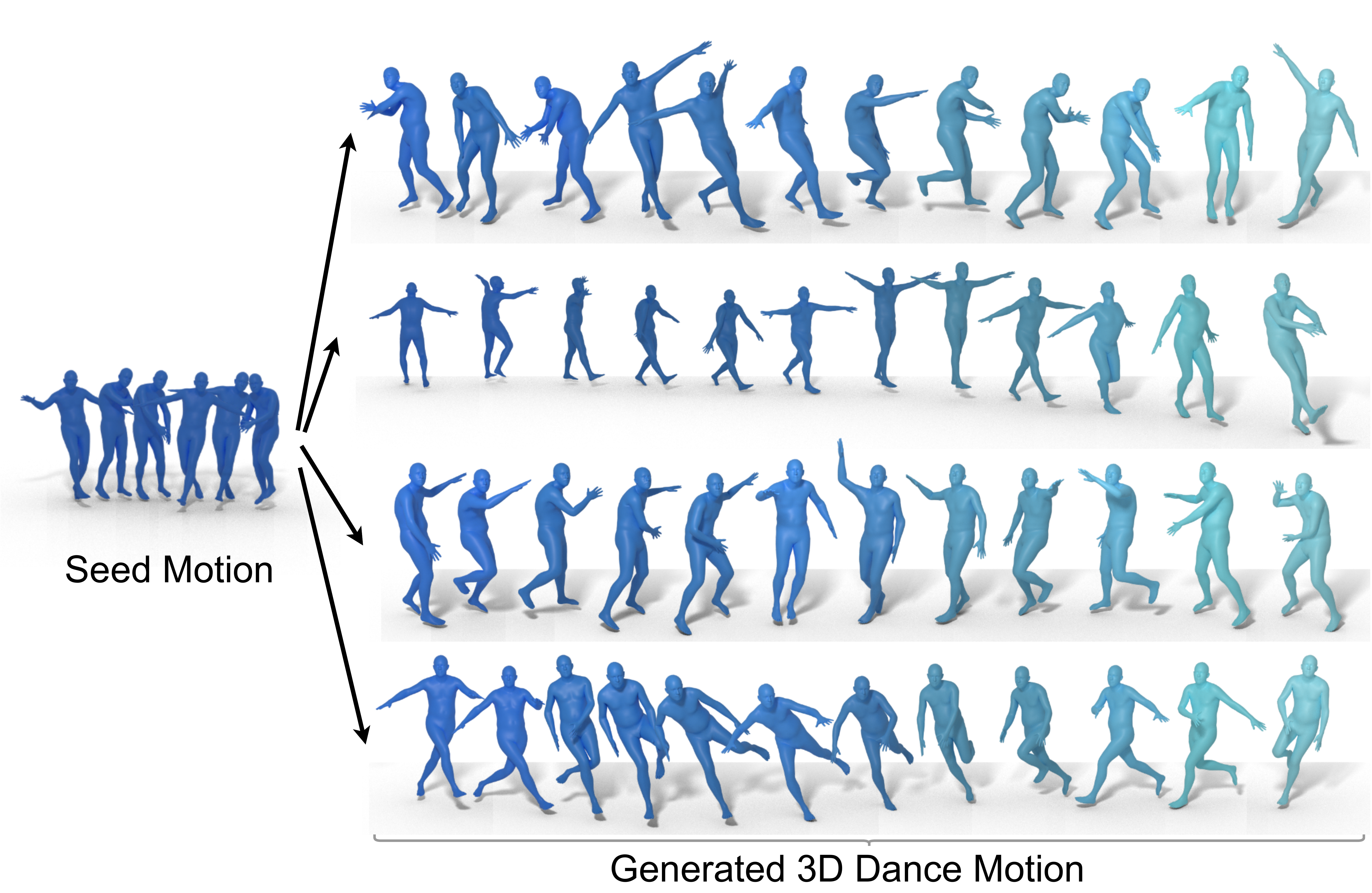}
\vspace{-1mm}
\caption{\textbf{Diverse Generation Results.} Here we visualize 4 different dance motions generated using \textit{different} music but the \textit{same} seed motion. 
On the left we illustrate the 2 second seed motion and on the right we show the generated 3D dance sequences subsampled by 2 seconds.  
For rows top to bottom, the genres of the conditioning music are: Break, Ballet Jazz, Krump and Middle Hip-hop. 
Note that the seed motion come from hip-hop dance.Our model is able to adapt the dance style when given a more modern dance music (second row: Ballet Jazz). Please see more results in the supplementary video. 
} 
\label{fig:diversity_vis}
\vspace{-5mm}
\end{figure}

\vspace{-3mm}
\paragraph{Generation Diversity}
We also evaluate our model's ability to generate {diverse dance motions when given various input music} compared with the baseline methods. Similar to the prior work~\cite{huang2021}, we calculate the {average Euclidean distance in the feature space across} $40$ generated motions on the AIST++ \emph{test} set  to measure the diversity. The motion diversity in the geometric feature space and in the kinetic feature space are noted as \Divm and \Divk, respectively.
Table~\ref{tab:exp_comparison} shows that our method generates more diverse dance motions comparing to the baselines {except Li~\etal~\cite{li2020learning}, which discretizes the motion, leading to discontinuous outputs that results in high \Divk.} 
Our generated diverse motions are visualized in Figure~\ref{fig:diversity_vis}.

\begin{figure}[t]
\vspace{-5mm}
\centering
\includegraphics[width=0.45\textwidth]{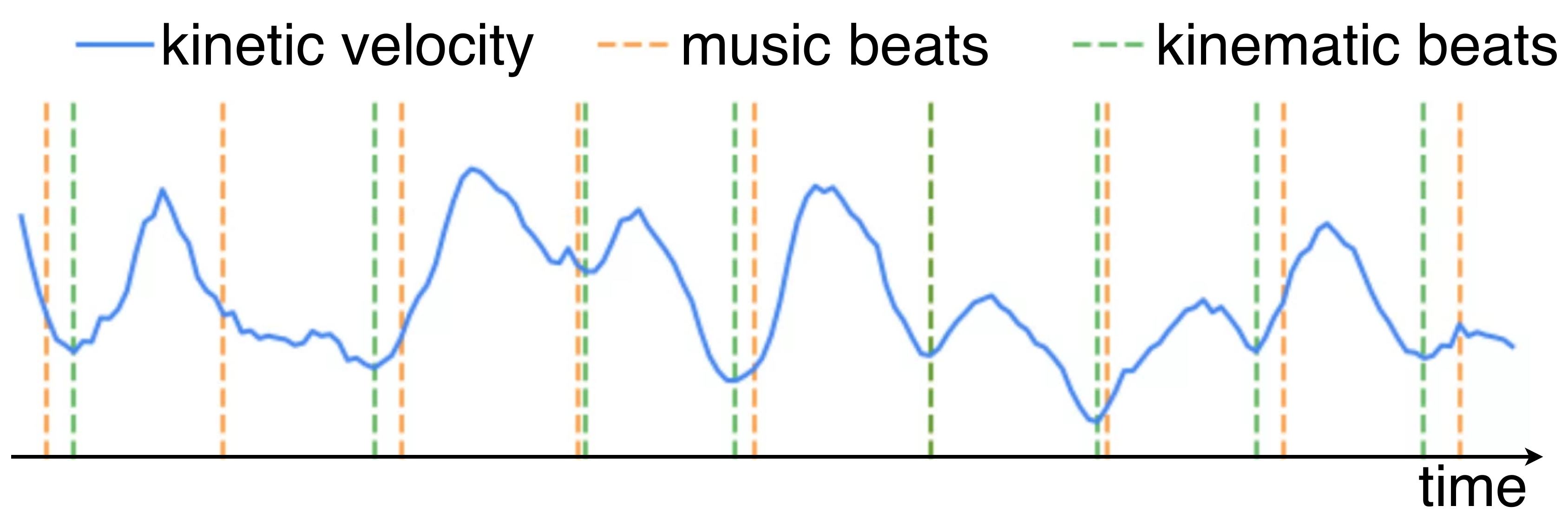}
\vspace{-3mm}
\caption{\textbf{Beats Alignment between Music and Generated Dance.} Here we visualize the kinetic velocity (blue curve) and kinematic beats (green dotted line) of our generated dance motion, as well as the music beats (orange dotted line). The kinematic beats are extracted by finding local minima from the kinetic velocity curve.
}
\label{fig:beat_curve}
\vspace{-5mm}
\end{figure}

\vspace{-3mm}
\paragraph{Motion-Music Correlation} Further, we evaluate how much the generated 3D motion correlates to the input music. 
As there is no well-designed metric to measure this property, we propose a novel metric, Beat Alignment Score (BeatAlign), to evaluate the motion-music correlation in terms of the similarity between the kinematic beats and music beats.
The music beats are extracted using \emph{librosa}~\cite{mcfee2015librosa} and the kinematic beats are computed as the local minima of the kinetic velocity, as shown in Figure~\ref{fig:beat_curve}.
The Beat Alignment Score is then defined as the average distance between every kinematic beat and its nearest music beat. 
Specifically, our Beat Alignment Score is defined as:
\begin{equation}
    \text{BeatAlign} = \frac{1}{m}\sum_{i=1}^{m}{\exp(-\frac{\min_{\forall t_j^y \in B^y} ||t_i^x - t_j^y||^2}{2\sigma^2})}
\end{equation}
where $B^x=\{t_i^x\}$ is the kinematic beats, $B^y = \{t_j^y\}$ is the music beats and $\sigma$ is a parameter to normalize sequences with different FPS. 
We set $\sigma = 3$ in all our experiments as the FPS of all our experiments sequences is 60.
A similar metric Beat Hit Rate was introduced in~\cite{lee2019dance, huang2021}, but this metric requires a dataset dependent handcrafted threshold to decide the alignment (``hit'') while ours directly measure the distances. 
This metric is explicitly designed to be uni-directional as dance motion does not necessarily \emph{have to} match with every music beat. On the other hand, every kinetic beat is expected to have a corresponding music beat.
To calibrate the results, we compute the correlation metrics on the entire AIST++ dataset (upper bound) and on the random-paired data (lower bound).
As shown in Table~\ref{tab:exp_comparison}, our generated motion is better correlated with the input music compared to the baselines. We also show one example in Figure~\ref{fig:beat_curve} that the kinematic beats of our generated motion align well with the music beats.
However, when comparing to the real data, all four methods including ours have a large space for improvement. This reflects that music-motion correlation is still a challenging problem.

\vspace{-3mm}
\subsubsection{Ablation Study}
\label{sec:ablation_study}
We conduct the following ablation experiments to study the effectiveness of our key design choices: Full-Attention Future-N supervision, and early cross-modal fusion. Please refer to our supplemental video for qualitative comparison.
The effectiveness of different model architectures is measured quantitatively using the motion quality (\FIDk, \FIDm) and the music-motion correlation (BeatAlign) metrics, as shown in Table~\ref{tab:exp_crossmodal} and Table~\ref{tab:exp_decoder}.

\vspace{-3mm}
\paragraph{Full-Attention Future-N Supervision}
\label{sec:attn_comp}
Here we dive deep into the attention mechanism and our future-N supervision scheme.
We set up four different settings: causal-attention shift-by-1 supervision, and full-attention with future-$\{1, 10, 20\}$ supervision.
Qualitatively, we find that the motion generated by the causal-attention with shift-by-1 supervision {(as done in \cite{li2020learning, radford2018improving,aksan2020attention})
starts to freeze after several seconds (please see the supplemental video).
Similar problem was reported in the results of ~\cite{aksan2020attention}.
Quantitatively (shown in the Table~\ref{tab:exp_decoder}), when using causal-attention shift-by-1 supervision, the FIDs are large meaning that the difference between generated and ground-truth motion sequences is substantial.} 
For the full-attention with future-1 supervision setting, the results rapidly drift during long-range generation. 
However, when the model is supervised with $10$ or $20$ future frames, it pays more attention to the temporal context. Thus, it learns to generate good quality (non-freezing, non-drifting) long-range motion.

\begin{table}[t]
\vspace{-5mm}
\centering{
\resizebox{1.0\linewidth}{!}{%
\midsepremove
\begin{tabular}[c]{lccc}
\toprule
Attn-Supervision & \FIDkArrow & \FIDmArrow & \BeatMetricArrow \\
\hline
Causal-Attn-Shift-by-1 & 111.69 & 21.43 & 0.217\\
Full-Attn-F1 (FACT-1) & 207.74 & 19.35 & 0.233\\
Full-Attn-F10 (FACT-10) & \textbf{35.10} & 15.17 & 0.239 \\
\hline
\rowcolor{Gray}
Full-Attn-F20 (FACT-20) & 35.35 & \textbf{12.39} & \textbf{0.241} \\
\bottomrule
\end{tabular}}}
\midsepdefault
\caption{\textbf{Ablation Study on Attention and Supervision Mechanism.} Causal-attention shift-by-1 supervision tends to generate freezing motions in the long-term. While Full-attention supervised more future frames boost the ability of generating more realistic dance motions. }
\label{tab:exp_decoder}
\vspace{-2mm}
\end{table}
\begin{figure}[t]
\centering
\includegraphics[width=0.45\textwidth]{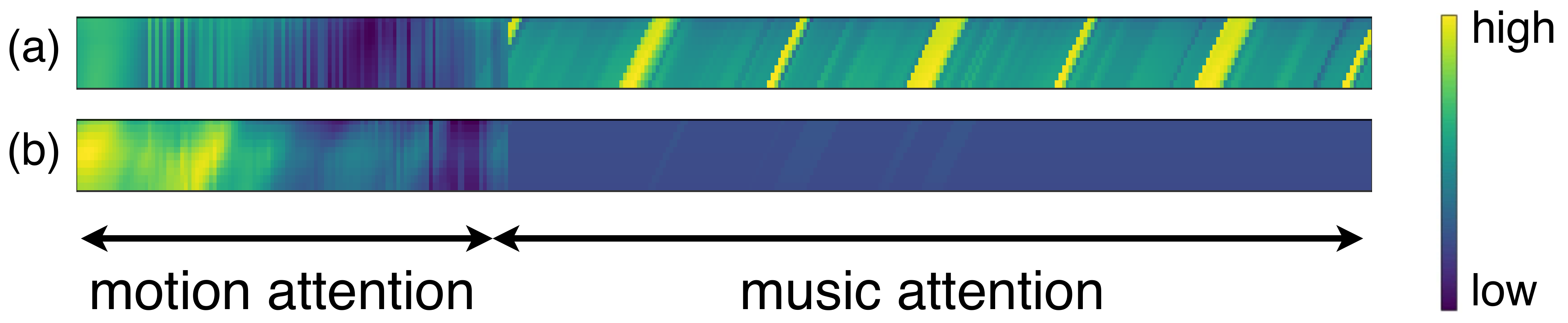}
\caption{\textbf{Attention Weights Visualization.} We compare the attention weights from the last layer of the (a) 12-layer cross-modal transformer and (b) 1-layer cross-modal transformer. Deeper cross-modal transformer pays equal attention to motion and music, while a shallower one pays more attention to motion.
}
\label{fig:atten_weights}
\vspace{-4mm}
\end{figure}

\vspace{-3mm}
\paragraph{Early Cross-Modal Fusion} 
Here we investigate when to fuse the two input modalities.
We conduct experiments in three settings, (1) \emph{No-Fusion}: 14-layer motion transformer only; (2) \emph{Late-Fusion}: 13-layer motion/audio transformer with 1-layer cross-modal transformer; (3) \emph{Early-Fusion}: 2-layer motion/audio transformer with 12-layer cross-modal transformer. 
For fair comparison, we change the number of attention layers in the motion/audio transformer and the cross-modal transformer but keep the total number of the attention layers fixed.
Table~\ref{tab:exp_crossmodal} shows that the early fusion between two input modalities is critical to generate motions that are well correlated with input music. Also we show in Figure~\ref{fig:atten_weights} that Early-Fusion allows the cross-model transformer pays more attention to the music, while Late-Fusion tend to ignore the conditioning music.
This also aligns with our intuition that the two modalities need to be fully fused for better cross-modal learning, as contrast to prior work that uses a single MLP to combine the audio and motion~\cite{li2020learning}.

\vspace{-3mm}
\subsubsection{User Study}
\label{sec:user_study}
Finally, we perceptually evaluate the motion-music correlation with a user study to compare our method with the three baseline methods and the ``random" baseline, which randomly combines AIST++ motion-music. (Refer to the Appendix for user study details.)
In this study, each user is asked to watch 10 videos showing one of our results and one random counterpart, and answer the question \emph{``which person is dancing more to the music? LEFT or RIGHT''} for each video. 
For user study on each of the four baselines, we invite 30 participants, ranging from professional dancers to people who rarely dance. 
We analyze the feedback and the results are: (1) $81\%$ of our generated dance motion is better than Li~\etal~\cite{li2020learning}; (2) $71\%$ of our generated dance motion is better than Dancenet~\cite{zhuang2020music2dance}; {(3) $77\%$ of our generated dance motion is better than DanceRevolution~\cite{huang2021}; }(4) $75\%$ of the unpaired AIST++ dance motion is better than ours. 
Clearly we surpass the baselines in the user study. 
But because the ``random'' baseline consists of real advanced dance motions that are extremely expressive, participants are biased to prefer it over ours. 
However, quantitative metrics show that our generated dance is more aligned with music.

\begin{table}[t]
\vspace{-5mm}
\centering{
\midsepremove
\begin{tabular}[c]{lccc}
\toprule
Cross-Modal Fusion & \FIDkArrow & \FIDmArrow & \BeatMetricArrow\\
\hline
No-Fusion & 45.66 & 13.27 & 0.228* \\
Late-Fusion  & 45.76 & 14.30 & 0.234 \\
\hline
\rowcolor{Gray}
Early-Fusion & \textbf{35.35} & \textbf{12.39} & \textbf{0.241} \\
\bottomrule
\end{tabular}}
\midsepdefault
\caption{{\textbf{Ablation Study on Cross-modal Fusion.} Early fusion of the two modalities allows the model to generate motion sequences align better with the conditioning music. *Note this number is calculated using the music paired with the input motion.}}
\label{tab:exp_crossmodal}
\vspace{-2mm}
\end{table}

\section{Conclusion and Discussion}
In this paper, we present a cross-modal transformer-based neural network architecture that can not only learn the audio-motion correspondence but also can generate non-freezing high quality 3D motion sequences conditioned on music. 
We also construct the largest 3D human dance dataset: AIST++. This proposed, multi-view, multi-genre, cross-modal 3D motion dataset can not only help research in the conditional 3D motion generation research but also human understanding research in general. 
While our results shows a promising direction in this problem of music conditioned 3D motion generation, there are more to be explored. First, our approach is kinematic based and we do not reason about physical interactions between the dancer and the floor. Therefore the global translation can lead to artifacts such as foot sliding and floating. Second, our model is currently deterministic. 
Exploring how to generate multiple realistic dance per music is an exciting direction.

\section{Acknowledgement}
We thank Chen Sun, Austin Myers, Bryan Seybold and Abhijit Kundu for helpful discussions. We thank Emre Aksan and Jiaman Li for sharing their code. We also thank Kevin Murphy for the early attempts on this direction, as well as Peggy Chi and Pan Chen for the help on user study experiments. 

{\small
\bibliographystyle{ieee_fullname}
\bibliography{egbib}

\begin{thebibliography}{10}\itemsep=-1pt

\bibitem{mixamo}
Mixamo.
\newblock \url{https://www.mixamo.com/}.

\bibitem{ahn2020generative}
Hyemin Ahn, Jaehun Kim, Kihyun Kim, and Songhwai Oh.
\newblock Generative autoregressive networks for 3d dancing move synthesis from
  music.
\newblock {\em IEEE Robotics and Automation Letters}, 5(2):3500--3507, 2020.

\bibitem{aksan2020attention}
Emre Aksan, Peng Cao, Manuel Kaufmann, and Otmar Hilliges.
\newblock Attention, please: A spatio-temporal transformer for 3d human motion
  prediction.
\newblock {\em arXiv preprint arXiv:2004.08692}, 2020.

\bibitem{aksan2019structured}
Emre Aksan, Manuel Kaufmann, and Otmar Hilliges.
\newblock Structured prediction helps 3d human motion modelling.
\newblock In {\em Proceedings of the IEEE International Conference on Computer
  Vision}, pages 7144--7153, 2019.

\bibitem{alemi2017groovenet}
Omid Alemi, Jules Fran{\c{c}}oise, and Philippe Pasquier.
\newblock Groovenet: Real-time music-driven dance movement generation using
  artificial neural networks.
\newblock {\em networks}, 8(17):26, 2017.

\bibitem{andriluka20142d}
Mykhaylo Andriluka, Leonid Pishchulin, Peter Gehler, and Bernt Schiele.
\newblock 2d human pose estimation: New benchmark and state of the art
  analysis.
\newblock In {\em Proceedings of the IEEE Conference on computer Vision and
  Pattern Recognition}, pages 3686--3693, 2014.

\bibitem{arikan2002interactive}
Okan Arikan and David~A Forsyth.
\newblock Interactive motion generation from examples.
\newblock {\em ACM Transactions on Graphics (TOG)}, 21(3):483--490, 2002.

\bibitem{bengio2015scheduled}
Samy Bengio, Oriol Vinyals, Navdeep Jaitly, and Noam Shazeer.
\newblock Scheduled sampling for sequence prediction with recurrent neural
  networks.
\newblock In {\em Advances in neural information processing systems}, 2015.

\bibitem{bhattacharya2021text2gestures}
Uttaran Bhattacharya, Nicholas Rewkowski, Abhishek Banerjee, Pooja Guhan,
  Aniket Bera, and Dinesh Manocha.
\newblock Text2gestures: A transformer-based network for generating emotive
  body gestures for virtual agents.
\newblock {\em arXiv preprint arXiv:2101.11101}, 2021.

\bibitem{bowden2000learning}
Richard Bowden.
\newblock Learning statistical models of human motion.
\newblock In {\em IEEE Workshop on Human Modeling, Analysis and Synthesis,
  CVPR}, volume 2000, 2000.

\bibitem{brand2000style}
Matthew Brand and Aaron Hertzmann.
\newblock Style machines.
\newblock In {\em Proceedings of the 27th annual conference on Computer
  graphics and interactive techniques}, pages 183--192, 2000.

\bibitem{butepage2017deep}
Judith B{\"u}tepage, Michael~J Black, Danica Kragic, and Hedvig Kjellstr{\"o}m.
\newblock Deep representation learning for human motion prediction and
  classification.
\newblock In {\em CVPR}, page 2017, 2017.

\bibitem{bybaby}
Slides by Saheel.
\newblock Baby talk: Understanding and generating image descriptions.

\bibitem{cao2018openpose}
Zhe Cao, Gines Hidalgo, Tomas Simon, Shih-En Wei, and Yaser Sheikh.
\newblock Open{P}ose: realtime multi-person 2{D} pose estimation using {P}art
  {A}ffinity {F}ields.
\newblock In {\em arXiv preprint arXiv:1812.08008}, 2018.

\bibitem{chen2021choreomaster}
Kang Chen, Zhipeng Tan, Jin Lei, Song-Hai Zhang, Yuan-Chen Guo, Weidong Zhang,
  and Shi-Min Hu.
\newblock Choreomaster: choreography-oriented music-driven dance synthesis.
\newblock {\em ACM Transactions on Graphics (TOG)}, 40(4):1--13, 2021.

\bibitem{chiu2019action}
Hsu-kuang Chiu, Ehsan Adeli, Borui Wang, De-An Huang, and Juan~Carlos Niebles.
\newblock Action-agnostic human pose forecasting.
\newblock In {\em 2019 IEEE Winter Conference on Applications of Computer
  Vision (WACV)}, pages 1423--1432. IEEE, 2019.

\bibitem{devlin2018bert}
Jacob Devlin, Ming-Wei Chang, Kenton Lee, and Kristina Toutanova.
\newblock Bert: Pre-training of deep bidirectional transformers for language
  understanding.
\newblock {\em arXiv preprint arXiv:1810.04805}, 2018.

\bibitem{du2019bio}
Xiaoxiao Du, Ram Vasudevan, and Matthew Johnson-Roberson.
\newblock Bio-lstm: A biomechanically inspired recurrent neural network for 3-d
  pedestrian pose and gait prediction.
\newblock {\em IEEE Robotics and Automation Letters}, 4(2):1501--1508, 2019.

\bibitem{duan2018weakly}
Xuguang Duan, Wenbing Huang, Chuang Gan, Jingdong Wang, Wenwu Zhu, and Junzhou
  Huang.
\newblock Weakly supervised dense event captioning in videos.
\newblock In {\em Advances in Neural Information Processing Systems}, pages
  3059--3069, 2018.

\bibitem{fan2011example}
Rukun Fan, Songhua Xu, and Weidong Geng.
\newblock Example-based automatic music-driven conventional dance motion
  synthesis.
\newblock {\em IEEE transactions on visualization and computer graphics},
  18(3):501--515, 2011.

\bibitem{ferreira2020learning}
Joao~P Ferreira, Thiago~M Coutinho, Thiago~L Gomes, Jos{\'e}~F Neto, Rafael
  Azevedo, Renato Martins, and Erickson~R Nascimento.
\newblock Learning to dance: A graph convolutional adversarial network to
  generate realistic dance motions from audio.
\newblock {\em Computers \& Graphics}, 94:11--21.

\bibitem{ferstl2018investigating}
Ylva Ferstl and Rachel McDonnell.
\newblock Investigating the use of recurrent motion modelling for speech
  gesture generation.
\newblock In {\em Proceedings of the 18th International Conference on
  Intelligent Virtual Agents}, pages 93--98, 2018.

\bibitem{ferstl2020adversarial}
Ylva Ferstl, Michael Neff, and Rachel McDonnell.
\newblock Adversarial gesture generation with realistic gesture phasing.
\newblock {\em Computers \& Graphics}, 2020.

\bibitem{fragkiadaki2015recurrent}
Katerina Fragkiadaki, Sergey Levine, Panna Felsen, and Jitendra Malik.
\newblock Recurrent network models for human dynamics.
\newblock In {\em Proceedings of the IEEE International Conference on Computer
  Vision}, pages 4346--4354, 2015.

\bibitem{galata2001learning}
Aphrodite Galata, Neil Johnson, and David Hogg.
\newblock Learning variable-length markov models of behavior.
\newblock {\em Computer Vision and Image Understanding}, 81(3):398--413, 2001.

\bibitem{gan2020foley}
Chuang Gan, Deng Huang, Peihao Chen, Joshua~B Tenenbaum, and Antonio Torralba.
\newblock Foley music: Learning to generate music from videos.
\newblock {\em arXiv preprint arXiv:2007.10984}, 2020.

\bibitem{ghosh2017learning}
Partha Ghosh, Jie Song, Emre Aksan, and Otmar Hilliges.
\newblock Learning human motion models for long-term predictions.
\newblock In {\em 2017 International Conference on 3D Vision (3DV)}, pages
  458--466. IEEE, 2017.

\bibitem{ginosar2019learning}
Shiry Ginosar, Amir Bar, Gefen Kohavi, Caroline Chan, Andrew Owens, and
  Jitendra Malik.
\newblock Learning individual styles of conversational gesture.
\newblock In {\em Proceedings of the IEEE Conference on Computer Vision and
  Pattern Recognition}, pages 3497--3506, 2019.

\bibitem{gopinath2020fairmotion}
Deepak Gopinath and Jungdam Won.
\newblock fairmotion - tools to load, process and visualize motion capture
  data.
\newblock Github, 2020.

\bibitem{heck2007parametric}
Rachel Heck and Michael Gleicher.
\newblock Parametric motion graphs.
\newblock In {\em Proceedings of the 2007 symposium on Interactive 3D graphics
  and games}, pages 129--136, 2007.

\bibitem{hernandez2019human}
Alejandro Hernandez, Jurgen Gall, and Francesc Moreno-Noguer.
\newblock Human motion prediction via spatio-temporal inpainting.
\newblock In {\em Proceedings of the IEEE International Conference on Computer
  Vision}, pages 7134--7143, 2019.

\bibitem{heusel2017gans}
Martin Heusel, Hubert Ramsauer, Thomas Unterthiner, Bernhard Nessler, and Sepp
  Hochreiter.
\newblock Gans trained by a two time-scale update rule converge to a local nash
  equilibrium.
\newblock In {\em Advances in neural information processing systems}, pages
  6626--6637, 2017.

\bibitem{holden2017phase}
Daniel Holden, Taku Komura, and Jun Saito.
\newblock Phase-functioned neural networks for character control.
\newblock {\em ACM Transactions on Graphics (TOG)}, 36(4):1--13, 2017.

\bibitem{holden2016deep}
Daniel Holden, Jun Saito, and Taku Komura.
\newblock A deep learning framework for character motion synthesis and editing.
\newblock {\em ACM Transactions on Graphics (TOG)}, 35(4):1--11, 2016.

\bibitem{holden2015learning}
Daniel Holden, Jun Saito, Taku Komura, and Thomas Joyce.
\newblock Learning motion manifolds with convolutional autoencoders.
\newblock In {\em SIGGRAPH Asia 2015 Technical Briefs}, pages 1--4. 2015.

\bibitem{huang2021}
Ruozi Huang, Huang Hu, Wei Wu, Kei Sawada, Mi Zhang, and Daxin Jiang.
\newblock Dance revolution: Long-term dance generation with music via
  curriculum learning.
\newblock In {\em International Conference on Learning Representations}, 2021.

\bibitem{DIP:SIGGRAPHAsia:2018}
Yinghao Huang, Manuel Kaufmann, Emre Aksan, Michael~J. Black, Otmar Hilliges,
  and Gerard Pons-Moll.
\newblock Deep inertial poser learning to reconstruct human pose from
  sparseinertial measurements in real time.
\newblock {\em ACM Transactions on Graphics, (Proc. SIGGRAPH Asia)},
  37(6):185:1--185:15, Nov. 2018.

\bibitem{iashin2020multi}
Vladimir Iashin and Esa Rahtu.
\newblock Multi-modal dense video captioning.
\newblock In {\em Proceedings of the IEEE/CVF Conference on Computer Vision and
  Pattern Recognition Workshops}, pages 958--959, 2020.

\bibitem{h36m_pami}
Catalin Ionescu, Dragos Papava, Vlad Olaru, and Cristian Sminchisescu.
\newblock Human3.6m: Large scale datasets and predictive methods for 3d human
  sensing in natural environments.
\newblock {\em IEEE Transactions on Pattern Analysis and Machine Intelligence},
  36(7):1325--1339, jul 2014.

\bibitem{jain2016structural}
Ashesh Jain, Amir~R Zamir, Silvio Savarese, and Ashutosh Saxena.
\newblock Structural-rnn: Deep learning on spatio-temporal graphs.
\newblock In {\em Proceedings of the ieee conference on computer vision and
  pattern recognition}, pages 5308--5317, 2016.

\bibitem{jia2019leveraging}
Ye Jia, Melvin Johnson, Wolfgang Macherey, Ron~J Weiss, Yuan Cao, Chung-Cheng
  Chiu, Naveen Ari, Stella Laurenzo, and Yonghui Wu.
\newblock Leveraging weakly supervised data to improve end-to-end
  speech-to-text translation.
\newblock In {\em ICASSP 2019-2019 IEEE International Conference on Acoustics,
  Speech and Signal Processing (ICASSP)}, pages 7180--7184. IEEE, 2019.

\bibitem{kao2020temporally}
Hsuan-Kai Kao and Li Su.
\newblock Temporally guided music-to-body-movement generation.
\newblock In {\em Proceedings of the 28th ACM International Conference on
  Multimedia}, pages 147--155, 2020.

\bibitem{karita2019comparative}
Shigeki Karita, Nanxin Chen, Tomoki Hayashi, Takaaki Hori, Hirofumi Inaguma,
  Ziyan Jiang, Masao Someki, Nelson Enrique~Yalta Soplin, Ryuichi Yamamoto,
  Xiaofei Wang, et~al.
\newblock A comparative study on transformer vs rnn in speech applications.
\newblock In {\em 2019 IEEE Automatic Speech Recognition and Understanding
  Workshop (ASRU)}, pages 449--456. IEEE, 2019.

\bibitem{karpathy2015deep}
Andrej Karpathy and Li Fei-Fei.
\newblock Deep visual-semantic alignments for generating image descriptions.
\newblock In {\em Proceedings of the IEEE conference on computer vision and
  pattern recognition}, pages 3128--3137, 2015.

\bibitem{kingma2014adam}
Diederik~P Kingma and Jimmy Ba.
\newblock Adam: A method for stochastic optimization.
\newblock {\em arXiv preprint arXiv:1412.6980}, 2014.

\bibitem{kocabas2019vibe}
Muhammed Kocabas, Nikos Athanasiou, and Michael~J. Black.
\newblock Vibe: Video inference for human body pose and shape estimation, 2019.

\bibitem{kovar2008motion}
Lucas Kovar, Michael Gleicher, and Fr{\'e}d{\'e}ric Pighin.
\newblock Motion graphs.
\newblock In {\em ACM SIGGRAPH 2008 classes}, pages 1--10. 2008.

\bibitem{krishna2017dense}
Ranjay Krishna, Kenji Hata, Frederic Ren, Li Fei-Fei, and Juan Carlos~Niebles.
\newblock Dense-captioning events in videos.
\newblock In {\em Proceedings of the IEEE international conference on computer
  vision}, pages 706--715, 2017.

\bibitem{kundu2020cross}
Jogendra~Nath Kundu, Himanshu Buckchash, Priyanka Mandikal, Anirudh Jamkhandi,
  Venkatesh~Babu RADHAKRISHNAN, et~al.
\newblock Cross-conditioned recurrent networks for long-term synthesis of
  inter-person human motion interactions.
\newblock In {\em Proceedings of the IEEE/CVF Winter Conference on Applications
  of Computer Vision}, pages 2724--2733, 2020.

\bibitem{todanceishuman}
Kimerer LaMothe.
\newblock The dancing species: how moving together in time helps make us human.
\newblock {\em Aeon}, June 2019.

\bibitem{lee2019dance}
Hsin-Ying Lee, Xiaodong Yang, Ming-Yu Liu, Ting-Chun Wang, Yu-Ding Lu,
  Ming-Hsuan Yang, and Jan Kautz.
\newblock Dancing to music, 2019.

\bibitem{lee2018listen}
Juheon Lee, Seohyun Kim, and Kyogu Lee.
\newblock Listen to dance: Music-driven choreography generation using
  autoregressive encoder-decoder network.
\newblock {\em arXiv preprint arXiv:1811.00818}, 2018.

\bibitem{lee1999hierarchical}
Jehee Lee and Sung~Yong Shin.
\newblock A hierarchical approach to interactive motion editing for human-like
  figures.
\newblock In {\em Proceedings of the 26th annual conference on Computer
  graphics and interactive techniques}, pages 39--48, 1999.

\bibitem{li2019entangled}
Guang Li, Linchao Zhu, Ping Liu, and Yi Yang.
\newblock Entangled transformer for image captioning.
\newblock In {\em Proceedings of the IEEE International Conference on Computer
  Vision}, pages 8928--8937, 2019.

\bibitem{li2020learning}
Jiaman Li, Yihang Yin, Hang Chu, Yi Zhou, Tingwu Wang, Sanja Fidler, and Hao
  Li.
\newblock Learning to generate diverse dance motions with transformer.
\newblock {\em arXiv preprint arXiv:2008.08171}, 2020.

\bibitem{li2018auto}
Zimo Li, Yi Zhou, Shuangjiu Xiao, Chong He, Zeng Huang, and Hao Li.
\newblock Auto-conditioned recurrent networks for extended complex human motion
  synthesis.
\newblock {\em ICLR}, 2018.

\bibitem{SMPL}
Matthew Loper, Naureen Mahmood, Javier Romero, Gerard Pons-Moll, and Michael~J.
  Black.
\newblock {SMPL}: A skinned multi-person linear model.
\newblock {\em SIGGRAPH Asia}, 2015.

\bibitem{lu2018neural}
Jiasen Lu, Jianwei Yang, Dhruv Batra, and Devi Parikh.
\newblock Neural baby talk.
\newblock In {\em Proceedings of the IEEE conference on computer vision and
  pattern recognition}, pages 7219--7228, 2018.

\bibitem{mahmood2019amass}
Naureen Mahmood, Nima Ghorbani, Nikolaus~F Troje, Gerard Pons-Moll, and
  Michael~J Black.
\newblock Amass: Archive of motion capture as surface shapes.
\newblock In {\em Proceedings of the IEEE International Conference on Computer
  Vision}, pages 5442--5451, 2019.

\bibitem{mcfee2015librosa}
Brian McFee, Colin Raffel, Dawen Liang, Daniel~PW Ellis, Matt McVicar, Eric
  Battenberg, and Oriol Nieto.
\newblock librosa: Audio and music signal analysis in python.
\newblock In {\em Proceedings of the 14th python in science conference},
  volume~8, 2015.

\bibitem{moelants2003dance}
Dirk Moelants.
\newblock Dance music, movement and tempo preferences.
\newblock In {\em Proceedings of the 5th Triennial ESCOM Conference}, pages
  649--652. Hanover University of Music and Drama, 2003.

\bibitem{muller2005efficient}
Meinard M{\"u}ller, Tido R{\"o}der, and Michael Clausen.
\newblock Efficient content-based retrieval of motion capture data.
\newblock In {\em ACM SIGGRAPH 2005 Papers}, pages 677--685. 2005.

\bibitem{onuma2008fmdistance}
Kensuke Onuma, Christos Faloutsos, and Jessica~K Hodgins.
\newblock Fmdistance: A fast and effective distance function for motion capture
  data.
\newblock In {\em Eurographics (Short Papers)}, pages 83--86, 2008.

\bibitem{pullen2000animating}
Katherine Pullen and Christoph Bregler.
\newblock Animating by multi-level sampling.
\newblock In {\em Proceedings Computer Animation 2000}, pages 36--42. IEEE,
  2000.

\bibitem{multiviewpose}
Haibo Qiu, Chunyu Wang, Jingdong Wang, Naiyan Wang, and Wenjun Zeng.
\newblock Cross view fusion for 3d human pose estimation.
\newblock In {\em International Conference on Computer Vision (ICCV)}, 2019.

\bibitem{radford2018improving}
Alec Radford, Karthik Narasimhan, Tim Salimans, and Ilya Sutskever.
\newblock Improving language understanding by generative pre-training, 2018.

\bibitem{ren2019music}
Xuanchi Ren, Haoran Li, Zijian Huang, and Qifeng Chen.
\newblock Music-oriented dance video synthesis with pose perceptual loss.
\newblock {\em arXiv preprint arXiv:1912.06606}, 2019.

\bibitem{ren2020self}
Xuanchi Ren, Haoran Li, Zijian Huang, and Qifeng Chen.
\newblock Self-supervised dance video synthesis conditioned on music.
\newblock In {\em Proceedings of the 28th ACM International Conference on
  Multimedia}, pages 46--54, 2020.

\bibitem{ren2019fastspeech}
Yi Ren, Yangjun Ruan, Xu Tan, Tao Qin, Sheng Zhao, Zhou Zhao, and Tie-Yan Liu.
\newblock Fastspeech: Fast, robust and controllable text to speech.
\newblock In {\em Advances in Neural Information Processing Systems}, pages
  3171--3180, 2019.

\bibitem{Ronchi_2017_ICCV}
Matteo~Ruggero Ronchi and Pietro Perona.
\newblock Benchmarking and error diagnosis in multi-instance pose estimation.
\newblock In {\em The IEEE International Conference on Computer Vision (ICCV)},
  Oct 2017.

\bibitem{shiratori2006dancing}
Takaaki Shiratori, Atsushi Nakazawa, and Katsushi Ikeuchi.
\newblock Dancing-to-music character animation.
\newblock In {\em Computer Graphics Forum}, volume~25, pages 449--458. Wiley
  Online Library, 2006.

\bibitem{shlizerman2018audio}
Eli Shlizerman, Lucio Dery, Hayden Schoen, and Ira Kemelmacher-Shlizerman.
\newblock Audio to body dynamics.
\newblock In {\em Proceedings of the IEEE Conference on Computer Vision and
  Pattern Recognition}, pages 7574--7583, 2018.

\bibitem{starke2019neural}
Sebastian Starke, He Zhang, Taku Komura, and Jun Saito.
\newblock Neural state machine for character-scene interactions.
\newblock {\em ACM Trans. Graph.}, 38(6):209--1, 2019.

\bibitem{starke2020local}
Sebastian Starke, Yiwei Zhao, Taku Komura, and Kazi Zaman.
\newblock Local motion phases for learning multi-contact character movements.
\newblock {\em ACM Transactions on Graphics (TOG)}, 39(4):54--1, 2020.

\bibitem{TopYTVideos}
Statista.
\newblock
  \url{https://www.statista.com/statistics/249396/top-youtube-videos-views/},
  2020.
\newblock Accessed: 2020-11-09.

\bibitem{sun2019learning}
Chen Sun, Fabien Baradel, Kevin Murphy, and Cordelia Schmid.
\newblock Learning video representations using contrastive bidirectional
  transformer.
\newblock {\em arXiv preprint arXiv:1906.05743}, 2019.

\bibitem{sun2019videobert}
Chen Sun, Austin Myers, Carl Vondrick, Kevin Murphy, and Cordelia Schmid.
\newblock Videobert: A joint model for video and language representation
  learning.
\newblock In {\em Proceedings of the IEEE International Conference on Computer
  Vision}, pages 7464--7473, 2019.

\bibitem{sun2020deepdance}
Guofei Sun, Yongkang Wong, Zhiyong Cheng, Mohan~S Kankanhalli, Weidong Geng,
  and Xiangdong Li.
\newblock Deepdance: Music-to-dance motion choreography with adversarial
  learning.
\newblock {\em IEEE Transactions on Multimedia}, 2020.

\bibitem{tang2018dance}
Taoran Tang, Jia Jia, and Hanyang Mao.
\newblock Dance with melody: An lstm-autoencoder approach to music-oriented
  dance synthesis.
\newblock In {\em Proceedings of the 26th ACM international conference on
  Multimedia}, pages 1598--1606, 2018.

\bibitem{taylor2009factored}
Graham~W Taylor and Geoffrey~E Hinton.
\newblock Factored conditional restricted boltzmann machines for modeling
  motion style.
\newblock In {\em Proceedings of the 26th annual international conference on
  machine learning}, pages 1025--1032, 2009.

\bibitem{tendulkar2020feel}
Purva Tendulkar, Abhishek Das, Aniruddha Kembhavi, and Devi Parikh.
\newblock Feel the music: Automatically generating a dance for an input song.
\newblock {\em arXiv preprint arXiv:2006.11905}, 2020.

\bibitem{aist-dance-db}
Shuhei Tsuchida, Satoru Fukayama, Masahiro Hamasaki, and Masataka Goto.
\newblock Aist dance video database: Multi-genre, multi-dancer, and
  multi-camera database for dance information processing.
\newblock In {\em Proceedings of the 20th International Society for Music
  Information Retrieval Conference, {ISMIR} 2019}, pages 501--510, Delft,
  Netherlands, Nov. 2019.

\bibitem{valin2019lpcnet}
Jean-Marc Valin and Jan Skoglund.
\newblock Lpcnet: Improving neural speech synthesis through linear prediction.
\newblock In {\em ICASSP 2019-2019 IEEE International Conference on Acoustics,
  Speech and Signal Processing (ICASSP)}, pages 5891--5895. IEEE, 2019.

\bibitem{vaswani2017attention}
Ashish Vaswani, Noam Shazeer, Niki Parmar, Jakob Uszkoreit, Llion Jones,
  Aidan~N Gomez, {\L}ukasz Kaiser, and Illia Polosukhin.
\newblock Attention is all you need.
\newblock In {\em Advances in neural information processing systems}, pages
  5998--6008, 2017.

\bibitem{venugopalan2015sequence}
Subhashini Venugopalan, Marcus Rohrbach, Jeffrey Donahue, Raymond Mooney,
  Trevor Darrell, and Kate Saenko.
\newblock Sequence to sequence-video to text.
\newblock In {\em Proceedings of the IEEE international conference on computer
  vision}, pages 4534--4542, 2015.

\bibitem{venugopalan2014translating}
Subhashini Venugopalan, Huijuan Xu, Jeff Donahue, Marcus Rohrbach, Raymond
  Mooney, and Kate Saenko.
\newblock Translating videos to natural language using deep recurrent neural
  networks.
\newblock {\em arXiv preprint arXiv:1412.4729}, 2014.

\bibitem{villegas2018neural}
Ruben Villegas, Jimei Yang, Duygu Ceylan, and Honglak Lee.
\newblock Neural kinematic networks for unsupervised motion retargetting.
\newblock In {\em CVPR}, 2018.

\bibitem{wang2019imitation}
Borui Wang, Ehsan Adeli, Hsu-kuang Chiu, De-An Huang, and Juan~Carlos Niebles.
\newblock Imitation learning for human pose prediction.
\newblock In {\em Proceedings of the IEEE International Conference on Computer
  Vision}, pages 7124--7133, 2019.

\bibitem{xu2020cross}
Haoming Xu, Runhao Zeng, Qingyao Wu, Mingkui Tan, and Chuang Gan.
\newblock Cross-modal relation-aware networks for audio-visual event
  localization.
\newblock In {\em Proceedings of the 28th ACM International Conference on
  Multimedia}, pages 3893--3901, 2020.

\bibitem{yalta2019weakly}
Nelson Yalta, Shinji Watanabe, Kazuhiro Nakadai, and Tetsuya Ogata.
\newblock Weakly-supervised deep recurrent neural networks for basic dance step
  generation.
\newblock In {\em 2019 International Joint Conference on Neural Networks
  (IJCNN)}, pages 1--8. IEEE, 2019.

\bibitem{yao2015describing}
Li Yao, Atousa Torabi, Kyunghyun Cho, Nicolas Ballas, Christopher Pal, Hugo
  Larochelle, and Aaron Courville.
\newblock Describing videos by exploiting temporal structure.
\newblock In {\em Proceedings of the IEEE international conference on computer
  vision}, pages 4507--4515, 2015.

\bibitem{ye2020choreonet}
Zijie Ye, Haozhe Wu, Jia Jia, Yaohua Bu, Wei Chen, Fanbo Meng, and Yanfeng
  Wang.
\newblock Choreonet: Towards music to dance synthesis with choreographic action
  unit.
\newblock In {\em Proceedings of the 28th ACM International Conference on
  Multimedia}, pages 744--752, 2020.

\bibitem{yu2016video}
Haonan Yu, Jiang Wang, Zhiheng Huang, Yi Yang, and Wei Xu.
\newblock Video paragraph captioning using hierarchical recurrent neural
  networks.
\newblock In {\em Proceedings of the IEEE conference on computer vision and
  pattern recognition}, pages 4584--4593, 2016.

\bibitem{zhang2018mode}
He Zhang, Sebastian Starke, Taku Komura, and Jun Saito.
\newblock Mode-adaptive neural networks for quadruped motion control.
\newblock {\em ACM Transactions on Graphics (TOG)}, 37(4):1--11, 2018.

\bibitem{zhou2018end}
Luowei Zhou, Yingbo Zhou, Jason~J Corso, Richard Socher, and Caiming Xiong.
\newblock End-to-end dense video captioning with masked transformer.
\newblock In {\em Proceedings of the IEEE Conference on Computer Vision and
  Pattern Recognition}, pages 8739--8748, 2018.

\bibitem{zhuang2020music2dance}
Wenlin Zhuang, Congyi Wang, Siyu Xia, Jinxiang Chai, and Yangang Wang.
\newblock Music2dance: Music-driven dance generation using wavenet.
\newblock {\em arXiv preprint arXiv:2002.03761}, 2020.

\bibitem{zhuang2020towards}
Wenlin Zhuang, Yangang Wang, Joseph Robinson, Congyi Wang, Ming Shao, Yun Fu,
  and Siyu Xia.
\newblock Towards 3d dance motion synthesis and control.
\newblock {\em arXiv preprint arXiv:2006.05743}, 2020.

\end{thebibliography}
}

\appendix
\section*{{\Large Appendix}}
\section{AIST++ Dataset Details}
\label{appendix:dataset}
\paragraph{3D Reconstruction} Here we describe how we reconstruct 3D motion from the AIST dataset. Although the AIST dataset contains multi-view videos, they are not calibrated meaning their camera intrinsic and extrinsic parameters are not available. Without camera parameters, it is not trivial to automatically and accurately reconstruct the 3D human motion. We start with 2D human pose detection~\cite{papandreou2017towards} and manually initialized the camera parameters. On this we apply bundle adjustment~\cite{triggs1999bundle} to refine the camera parameters. With the improved camera parameters, the 3D joint locations $\hat{J}\in\mathbb{R}^{M\times 3}$($M=17$) are then triangulated from the multi-view 2D human pose keypoints locations. 
During the triangulation phase, we introduce temporal smoothness and bone length constraints to improve the quality of the reconstructed 3D joint locations. 
We further fit SMPL human body model~\cite{loper2015smpl} to the triangulated joint locations $\hat J$ by minimizing an objective with respect to $\Theta = \{\theta_i\}_i^M$, global scale parameter $\alpha$ and global transformation 
$\gamma$ for each frame:
$    \min_{\Theta, \gamma,\alpha} \sum_{i=1}^{M}\|\hat{J}-J(\theta_i, \beta, \gamma,\alpha)\|_2$.
We fix $\beta$ to the average shape as the problem is under-constrained from 3D joint locations alone. 

\begin{figure}[t]
\centering
\includegraphics[width=0.45\textwidth]{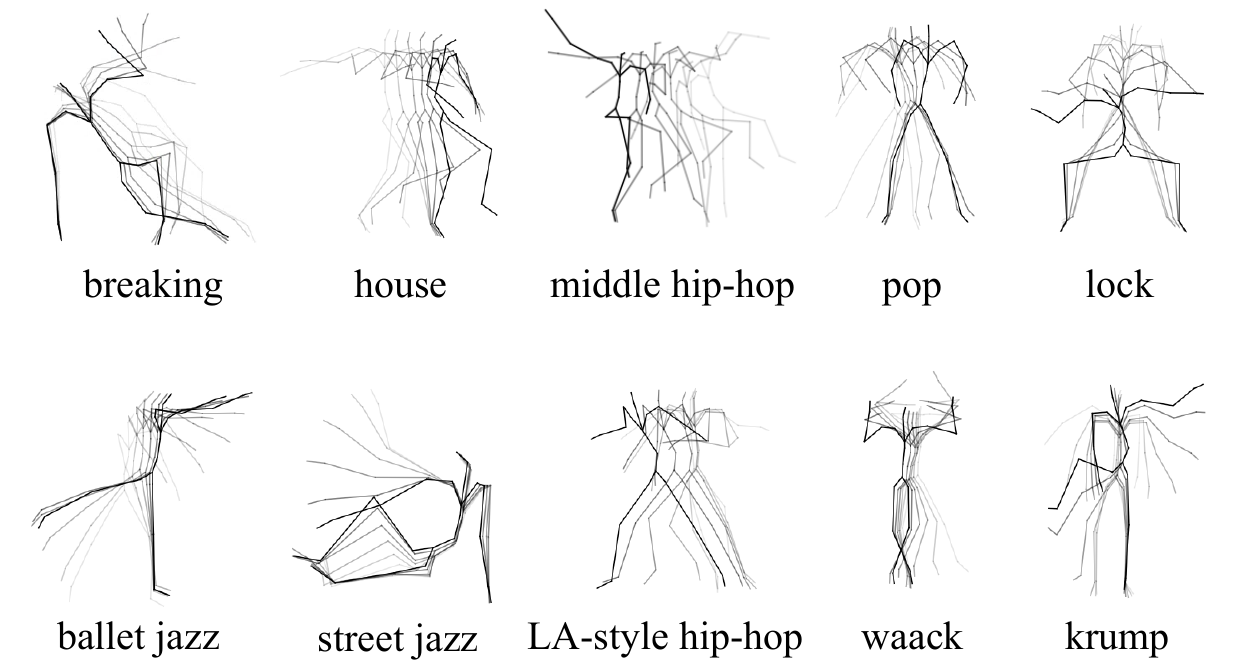}
\caption{\textbf{AIST++ Motion Diversity Visualization.} Here we show the 10 types of 3D human dance motion in our dataset.}
\label{fig:aist_viz}
\vspace{-3mm}
\end{figure}

\paragraph{Statistics} We show the detailed statistics of our AIST++ dataset in Table~\ref{tab:dataset_stats}. Thanks to the AIST Dance Video Database~\cite{aist-dance-db}, our dataset contains in total $5.2$-hour ($1.1$M frame, $1408$ sequences) of 3D dance motion accompanied with music. 
The dataset covers 10 dance genre (shown in Figure~\ref{fig:aist_viz}) and $60$ pieces of music. 
For each genre, there are $6$ different pieces of music, ranging from $29$ seconds to $54$ seconds long, and from $80$ BPM to $130$ BPM (except for House genre which is $110$ BPM to $135$ BPM).
Among those motion sequences for each genre, $120$ ($85$\%) of them are \emph{basic}  choreographies and $21$ ($15$\%) of them are \emph{advanced}.
Advanced choreographies are longer and more complicated dances improvised by the dancers. 
Note for the \emph{basic} dance motion, dancers are asked to perform the same choreography on all the $6$ pieces of music with different speed to follow different music BPMs. 
So the total \emph{unique} choreographies in for each genre is $120/6 + 21 = 41$. 
In our experiments we split the AIST++ dataset such that there is no overlap between  \emph{train} and \emph{test} for both music and choreographies (see Sec. 5.2.1 in the paper).

\begin{table*}[t]
\centering{
\resizebox{1.0\linewidth}{!}{%
\begin{tabular}[c]{ccccccc}
\hline
Genres & Musics & Music Tempo & Motions & Choreographs & Motion Duration (sec.) & Total Seconds  \\
\hline
ballet jazz & 6 & 80 - 130 & 141 & \multirow{10}{6em}{85\% basic + 15\% advanced} & 7.4 - 12.0 basic / 29.5 - 48.0 adv. & 1910.8 \\
street jazz & 6 & 80 - 130 & 141 &  & 7.4 - 12.0 basic / 14.9 - 48.0 adv. & 1875.3 \\
krump & 6 & 80 - 130 & 141 &  & 7.4 - 12.0 basic / 29.5 - 48.0 adv. & 1904.3 \\
house & 6 & 110 - 135 & 141 &  & 7.1 - 8.7 basic / 28.4 - 34.9 adv. & 1607.6 \\
LA-style hip-hop & 6 & 80 - 130 & 141 &  & 7.4 - 12.0 basic / 29.5 - 48.0 adv. & 1935.8 \\
middle hip-hop & 6 & 80 - 130 & 141 &  & 7.4 - 12.0 basic / 29.5 - 48.0 adv. & 1934.0 \\
waack & 6 & 80 - 130 & 140 &  & 7.4 - 12.0 basic / 29.5 - 48.0 adv. & 1897.1 \\
lock & 6 & 80 - 130 & 141 &  & 7.4 - 12.0 basic / 29.5 - 48.0 adv. & 1898.5 \\
pop & 6 & 80 - 130 & 140 &  & 7.4 - 12.0 basic / 29.5 - 48.0 adv. & 1872.9 \\
break & 6 & 80 - 130 & 141 &  & 7.4 - 12.0 basic / 23.8 - 48.0 adv. & 1858.3 \\
\hline
total & 60 & & 1408 &  &  & 18694.6
\end{tabular}}}
\caption{\textbf{AIST++ Dataset Statistics}. AIST++ is built upon a subset of AIST database~\cite{aist-dance-db} that contains single-person dance.}
\label{tab:dataset_stats}
\vspace{-3mm}
\end{table*}
\begin{figure}[t]
\centering
\includegraphics[width=0.45\textwidth]{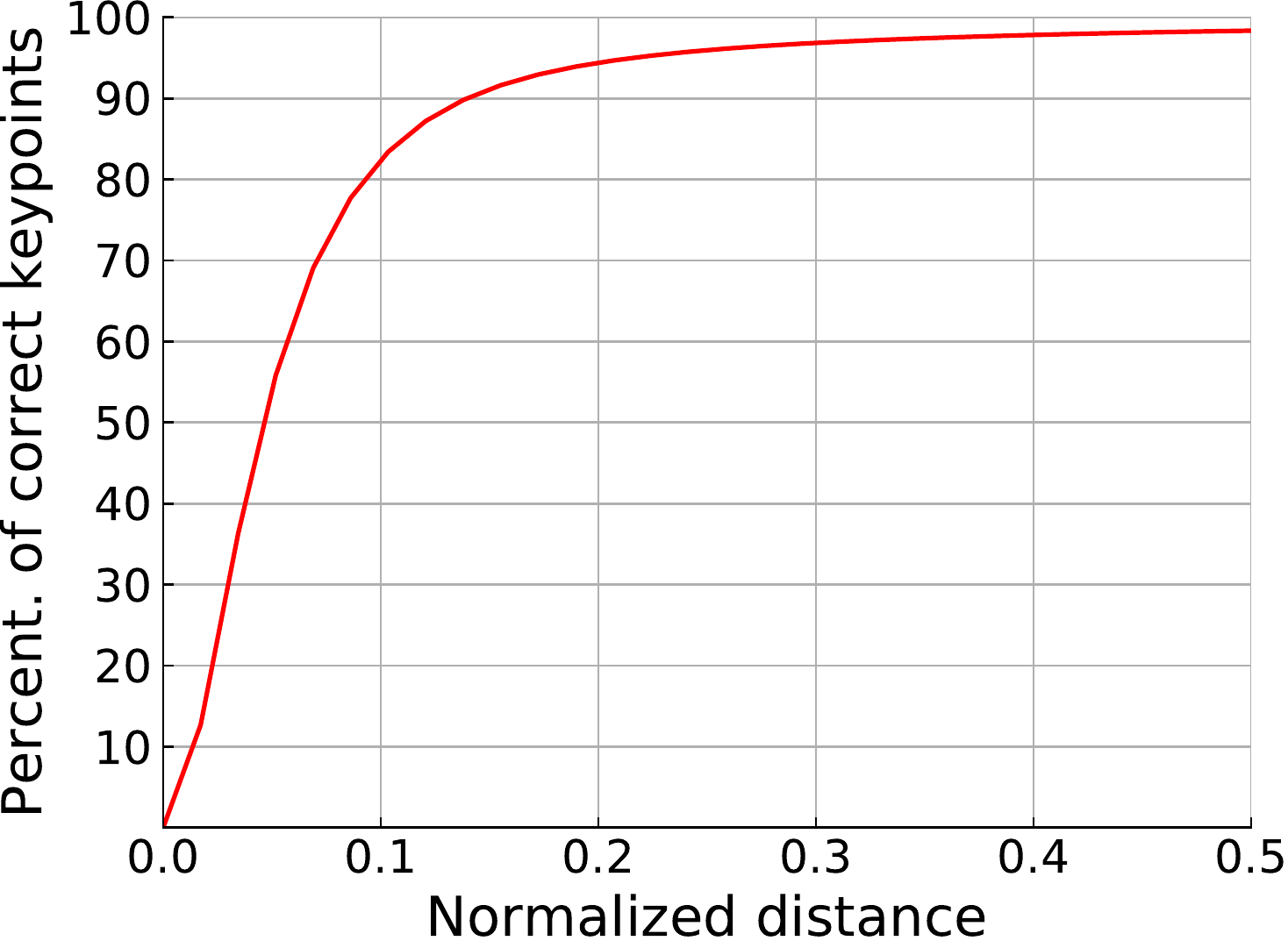}
\caption{\textbf{PCKh Metric on AIST++.} We analyze the PCKh (percentage of correct keypoints) metric between re-projected 2D keypoints and detected 2D keypoints on AIST++. Averaged PCKh@0.5 is $98.4\%$ on all joints shows that our reconstructed 3D keypoints are highly consistent with the predicted 2D keypoints.}
\label{fig:aist_pckh}
\vspace{-3mm}
\end{figure}
\begin{figure}[t]
\centering
\includegraphics[width=0.45\textwidth]{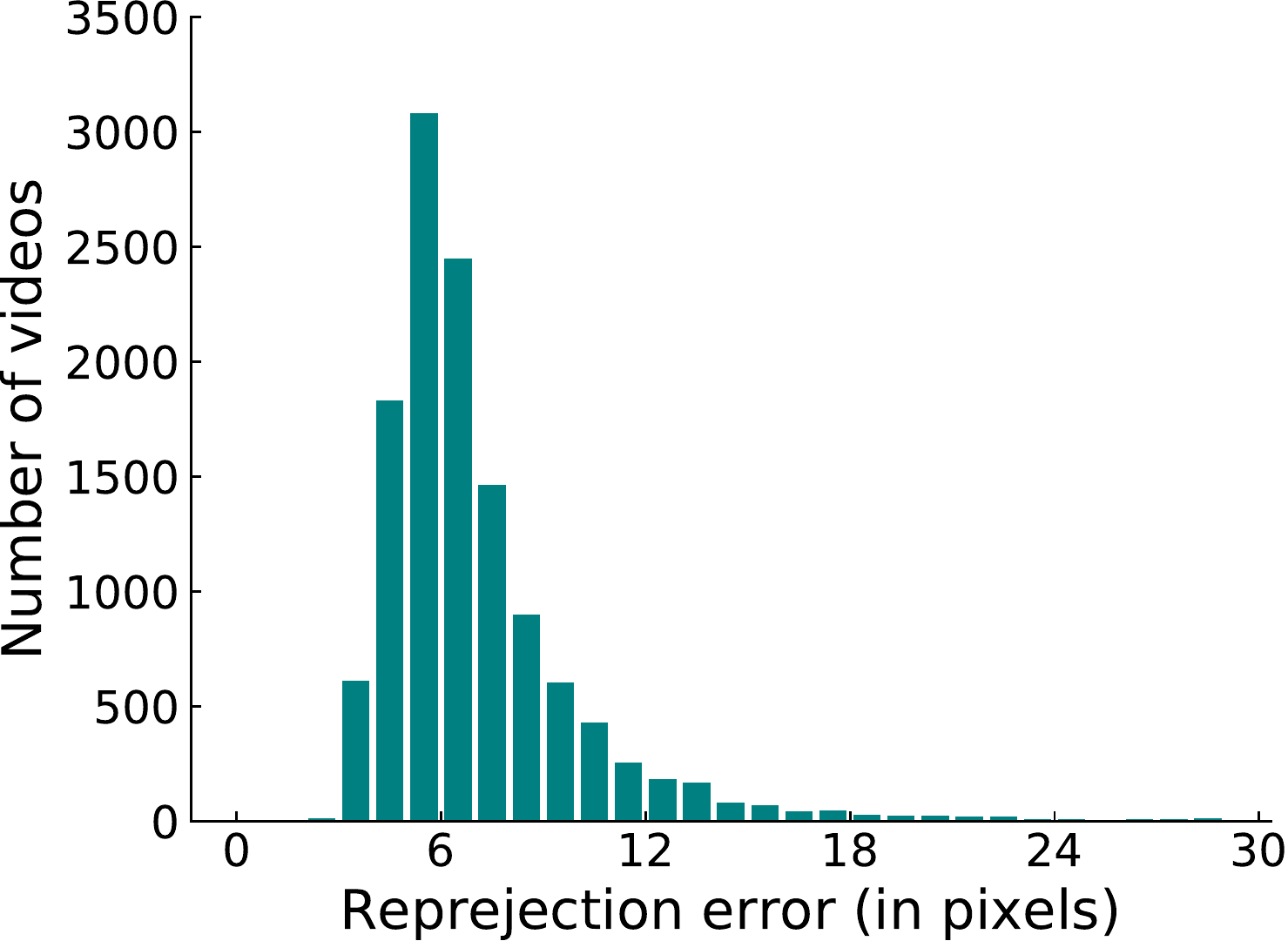}
\caption{\textbf{MPJPE-2D Distribution on AIST++.} We analyze the distribution of MPJPE-2D among all video sequences on 1920x1080 resolution. MPJPE-2D is calculated between the re-projected 2D keypoints and the detected 2D keypoints. Over $86\%$ of the videos have less than average $10$ pixels of error. }
\label{fig:aist_reproj_error}
\vspace{-3mm}
\end{figure}
\begin{figure*}[t]
\centering
\includegraphics[width=0.8\textwidth]{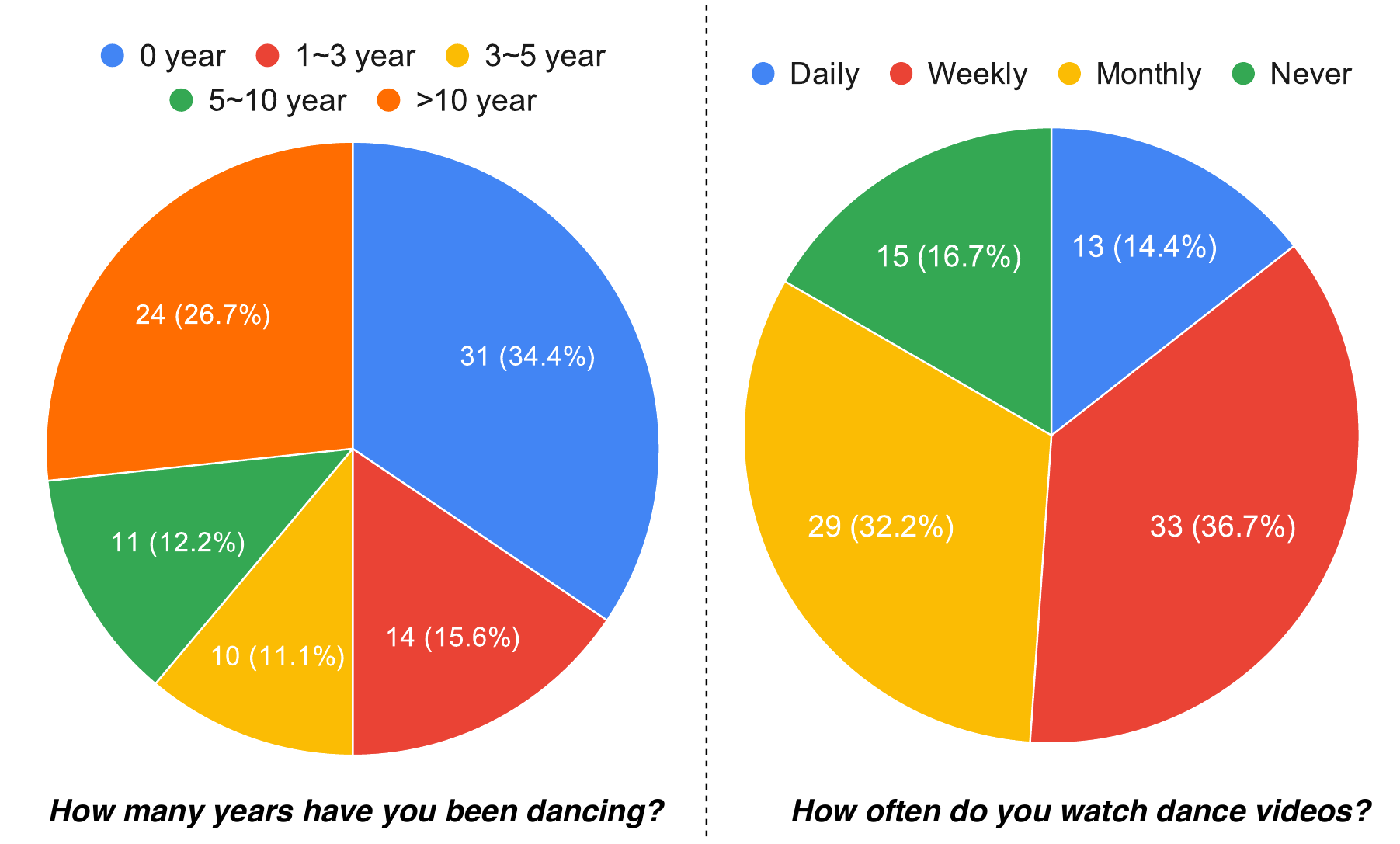}
\caption{\textbf{Participant Demography of the \emph{Comparison} User Study.}
}
\label{fig:user_study_1_years}
\vspace{-3mm}
\end{figure*}
\begin{figure*}[t]
\centering
\includegraphics[width=0.8\textwidth]{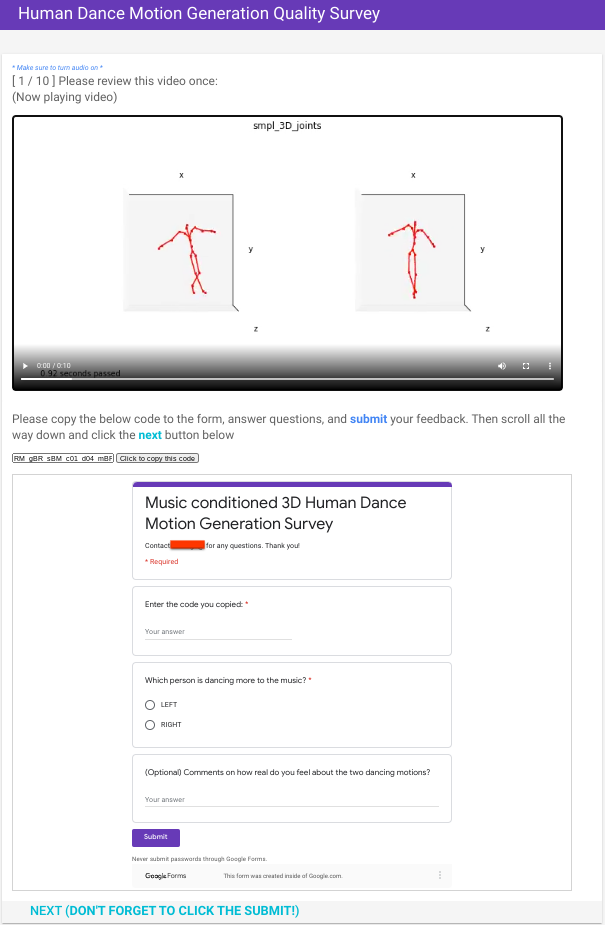}
\caption{\textbf{User study interface.} The interface of our User study. We ask each participant to watch 10 videos and answer the question \emph{"which person is dancing more to the music? LEFT or RIGHT"}.
}
\vspace{-3mm}
\label{fig:user_study_1_interface}
\end{figure*}

\paragraph{Validation} As described in Sec. 5.1 in the paper, we validate the quality of our reconstructed 3D motion by calculating the overall MPJPE-2D (in pixel) between the re-projected 2D keypoints and the detected 2D keypoints with high confidence ($> 0.5$). We provide here the distribution of MPJPE-2D among all video sequences (Figure~\ref{fig:aist_reproj_error}). Moreover, we also analyze the PCKh metric with various thresholds on the AIST++, which measures the consistency between the re-projected and detected 2D keypoints. Averaged PCKh@0.5 is 98.4\% on all joints shows that our reconstructed 3D keypoints are highly consistent with the detected 2D keypoints.

\section{User Study Details}

\label{appendix:userstudy}
\subsection{Comparison User Study}
As mentioned in Sec. 5.2.5 in the main paper, we qualitatively compare our generated results with several baselines in a user study. Here we describe the details of this user study. 
Figure~\ref{fig:user_study_1_interface} shows the interface that we developed for this user study. 
We visualize the dance motion using stick-man and conduct side-by-side comparison between our generated results and the baseline methods.
The left-right order is randomly shuffled for each video to make sure that the participants have absolutely no idea which is ours. 
Each video is $10$-second long, accompanied with the music. 
The question we ask each participant is \emph{``which person is dancing more to the music? LEFT or RIGHT''}, and the answers are collected through a Google Form. 
At the end of this user study, we also have an exit survey to ask for the dance experience of the participants. There are two questions: \emph{``How many years have you been dancing?''}, and \emph{``How often do you watch dance videos?''}. Figure~\ref{fig:user_study_1_years} shows that our participants ranges from professional dancers to people rarely dance, with majority with at least 1 year of dance experience.

\end{document}